\documentclass{article}

\PassOptionsToPackage{numbers, compress}{natbib}
\usepackage[final]{neurips_2024}

\usepackage[utf8]{inputenc} 
\usepackage[T1]{fontenc}    
\usepackage{hyperref}       
\usepackage{url}            
\usepackage{booktabs}       
\usepackage{amsfonts}       
\usepackage{nicefrac}       
\usepackage{microtype}      

\usepackage[dvipsnames]{xcolor}

\usepackage{graphicx}
\usepackage{cuted}
\usepackage{capt-of}
\usepackage{amsmath}
\usepackage{caption}
\usepackage{subcaption}
\usepackage{amssymb}
\usepackage{multibib}
\usepackage{wrapfig}
\usepackage{multirow}
\usepackage{multicol}
\usepackage{booktabs}
\usepackage{enumitem} 
\usepackage{pifont}
\usepackage{lipsum}
\usepackage{float}
\usepackage{array}

\newcommand{\cmark}{\textcolor{ForestGreen}{\ding{51}}}
\newcommand{\xmark}{\textcolor{Red}{\ding{55}}}

\renewcommand{\paragraph}[1]{{\parindent0pt \textbf{#1.}}}

\newcommand{\modelname}{Meta-Controller}

\newcolumntype{R}{c!{\vrule width 0.75pt}}

\title{\modelname{}: Few-Shot Imitation of Unseen Embodiments and Tasks in Continuous Control}

\author{
    Seongwoong Cho\thanks{Equal contribution}\hspace{1em}
    Donggyun Kim\footnotemark[1]\hspace{1em}
    Jinwoo Lee\hspace{1em}
    Seunghoon Hong\hspace{1em} \\
    School of Computing, KAIST \\
    \texttt{\{seongwoongjo, kdgyun425, bestgenius10, seunghoon.hong\}@kaist.ac.kr}    
}

\begin{document}

\maketitle

\begin{abstract}
\vspace{-0.3cm}
Generalizing across robot embodiments and tasks is crucial for adaptive robotic systems.
Modular policy learning approaches adapt to new embodiments but are limited to specific tasks, while few-shot imitation learning (IL) approaches often focus on a single embodiment.
In this paper, we introduce a few-shot behavior cloning framework to simultaneously generalize to unseen embodiments and tasks using a few (\emph{e.g.,} five) reward-free demonstrations.
Our framework leverages a joint-level input-output representation to unify the state and action spaces of heterogeneous embodiments and employs a novel structure-motion state encoder that is parameterized to capture both shared knowledge across all embodiments and embodiment-specific knowledge.
A matching-based policy network then predicts actions from a few demonstrations, producing an adaptive policy that is robust to over-fitting.
Evaluated in the DeepMind Control suite, our framework termed \modelname{} demonstrates superior few-shot generalization to unseen embodiments and tasks over modular policy learning and few-shot IL approaches.
Codes are available at \href{https://github.com/SeongwoongCho/meta-controller}{https://github.com/SeongwoongCho/meta-controller}.
\end{abstract}

\section{Introduction}
\label{sec:intro}

Generalizing across different robot embodiments and tasks with only a few demonstrations is a fundamental challenge in continuous control~\cite{devin2017learning,ghadirzadeh2021bayesian,zhou2022modularity,schubert2023generalist,vuong2023open,hansen2023td}.
This capability is crucial for developing versatile and adaptive robotic systems that can operate effectively in diverse and dynamic environments.
The high diversity of embodiments and tasks, however, makes this particularly challenging.
Robot embodiments vary widely in their morphological (\emph{e.g.,} number and connectivity of joints) and dynamic (\emph{e.g.,} motor gear ratios, damping coefficients) configurations, complicating the design of a unified architecture capable of handling heterogeneous input states and output actions.
Furthermore, the variety of tasks—such as locomotion, object manipulation, and navigation—requires the learning of transferable skills that can efficiently generalize across different tasks.

Despite significant advancements in reinforcement learning (RL) and imitation learning (IL), achieving simultaneous generalization across diverse embodiments and tasks with a few demonstrations remains largely underexplored.
Modular policy learning approaches~\cite{wang2018nervenet,huang2020one,hong2021structure,gupta2021metamorph,furuta2022system} have shown promise by learning modular policies that can be shared across embodiments with different morphologies.
However, these methods primarily focus on specific task types such as locomotion, and lack the flexibility to adapt to a wide range of control tasks, limiting their broader applicability.
Conversely, few-shot IL approaches~\cite{finn2017model,duan2017one,yoon2018bayesian,hakhamaneshi2021hierarchical,xu2022prompting,xu2022hyper} aim to learn novel tasks from a few demonstrations. These techniques excel in scenarios with sparse training data but typically concentrate on a single embodiment with fixed morphological and dynamic structures, restricting their ability to generalize across various embodiments and tasks.
As a result, these two fields have developed independently, and their integration remains an open area of research.

To address these challenges, we propose a novel framework that flexibly learns arbitrary control tasks on unseen embodiments using a few (\emph{e.g.,} five) reward-free demonstrations.
To handle heterogeneous embodiments within a unified architecture, we tokenize states and actions into joint-level representations, since joints serve as the fundamental building blocks of robots and provide a modular representation for compositional generalization to unseen embodiments~\cite{gupta2021metamorph,furuta2022system}.
Given this unified I/O, we employ a state encoder to capture both embodiment-specific knowledge about morphology and dynamics and shared knowledge about the physics governing the environment. 
Then we design a matching-based policy network that predicts actions from the encoded states, conditioned on a few demonstrations.
Our model is trained using episodic meta-learning on a dataset comprising various embodiments and tasks, followed by few-shot behavior cloning on unseen embodiments and tasks using a few reward-free demonstrations.

Our key contributions are as follows:
(1) We design a novel structure-motion encoder that operates on joint-level state representations, efficiently disentangling embodiment-specific and task-specific knowledge.
(2) We propose a matching-based meta-learning framework that efficiently transfers knowledge of local motions to quickly learn unseen tasks with a few demonstrations.
(3) We evaluate our framework in various environments within the DeepMind Control suite, encompassing diverse embodiments and tasks, demonstrating superior few-shot learning performance over existing baselines.
\section{Problem Setup}
\label{sec:problem_setup}

A reinforcement learning (RL) problem involves an agent interacting with an environment, typically modeled as a Markov Decision Process (MDP).
An MDP is represented by the tuple $(\mathcal{S}, \mathcal{A}, P, R)$, where $\mathcal{S}$ is the state space, $\mathcal{A}$ is the action space, $P: \mathcal{S} \times \mathcal{A} \times \mathcal{S} \to \mathbb{R}$ is the transition probability, 
and $R: \mathcal{S} \times \mathcal{A} \xrightarrow{} \mathbb{R}$ is the reward function.
In conventional RL, an agent learns a policy $\pi: \mathcal{S} \xrightarrow{} \mathcal{A}$ that maximizes expected cumulative rewards $\mathbb{E}_{\pi} [\gamma^t R(s_t, a_t)]$, where $\gamma^t \in [0, 1]$ is a discount factor.
Training such an RL agent typically requires numerous interactions with the environment and a carefully designed reward function, making it burdensome to learn new tasks.
Behavior cloning (BC) addresses these challenges by using supervised learning techniques to imitate an expert policy from offline demonstrations.
We focus on the few-shot BC setting, where the training data consists of a few reward-free demonstrations $\mathcal{D} = \{\tau_i\}_{i \leq N}$, with each demonstration $\tau_i = \{(s_t^i, a_t^i)\}_{t \le T}$ being a temporal sequence of states and actions performed by an expert model.

In this paper, we consider continuous control problems of multi-joint robots that involve various embodiments and tasks.
An embodiment $\mathcal{E}$ refers to the physical configuration of robots, which includes (1) the morphology, \emph{i.e.,} the shape, size, and arrangement of the components such as limbs, joints, and sensors, and (2) the dynamics parameters that affect the robot's behavior, \emph{e.g.,} motion inertia, mass, gear ratios, and damping.
In terms of MDP, different embodiments can have different dimensionality of state and action spaces, \emph{e.g.}, $\dim(\mathcal{S}_{\mathcal{E}_i}) \ne \dim(\mathcal{S}_{\mathcal{E}_j})$ and $\dim(\mathcal{A}_{\mathcal{E}_i}) \ne \dim(\mathcal{A}_{\mathcal{E}_j})$ for $\mathcal{E}_i \ne \mathcal{E}_j$, as well as distinct transition probabilities $P_\mathcal{E}$ that determine the kinematics of the robot.
A task $\mathcal{T}$ is defined by a specific goal or objective that the robot must achieve within its environment, characterized by a reward function $R_\mathcal{T}$.
Tasks can vary widely, ranging from locomotion and manipulation to complex interactions with dynamic environments.
The combination of different embodiments $\mathcal{E}$ and tasks $\mathcal{T}$ creates a broad class of continuous control problems.

Our objective is to achieve simultaneous generalization to unseen embodiments and tasks of continuous control with a few-shot behavior cloning framework.
In other words, the model has to learn a policy for a novel continuous control problem from only a few demonstrations $\mathcal{D}$, where both embodiment $\mathcal{E}$ and task $\mathcal{T}$ can be arbitrary and previously unseen.


\subsection{Challenges and Desiderata}
\label{sec:challenges_and_desiderata}
Despite achieving the simultaneous few-shot generalization to unseen embodiments and tasks is crucial for developing versatile and adaptive robotic systems, this problem remains less explored.
We characterize two distinct challenges and desiderata to address each challenge.

\paragraph{Handling Heterogeneous Embodiments}
To generalize to arbitrary embodiments in continuous control, the model must possess an architecture capable of universally handling heterogeneous states and actions of various embodiments.
This necessitates a unified input/output (I/O) representation for states and actions, allowing for the sharing of structural characteristics (e.g., dimensionality) and semantics (e.g., input attributes or output control types) across different embodiments.
Additionally, the encoder to extract state features should be able to capture transferable knowledge across different embodiments as well as embodiment-specific knowledge to flexibly adapt to distinct morphologies and dynamics of each embodiment.


\paragraph{Few-shot Policy Adaptation}
To achieve robust few-shot learning on unseen tasks, an efficient and flexible policy adaptation mechanism is essential.
Given the diverse behaviors required by continuous control problems across various embodiments and tasks, the policy network must learn transferable skills shared by different tasks.
Simultaneously, the model must dynamically adapt its policy by inferring the task based on a few demonstrations, ensuring it captures the underlying structure of previously unseen tasks. Additionally, the adaptation mechanism must be robust to overfitting.





\section{Method}
\label{sec:method}

\begin{figure}[t!]
    \centering
    \includegraphics[width=0.8\linewidth]{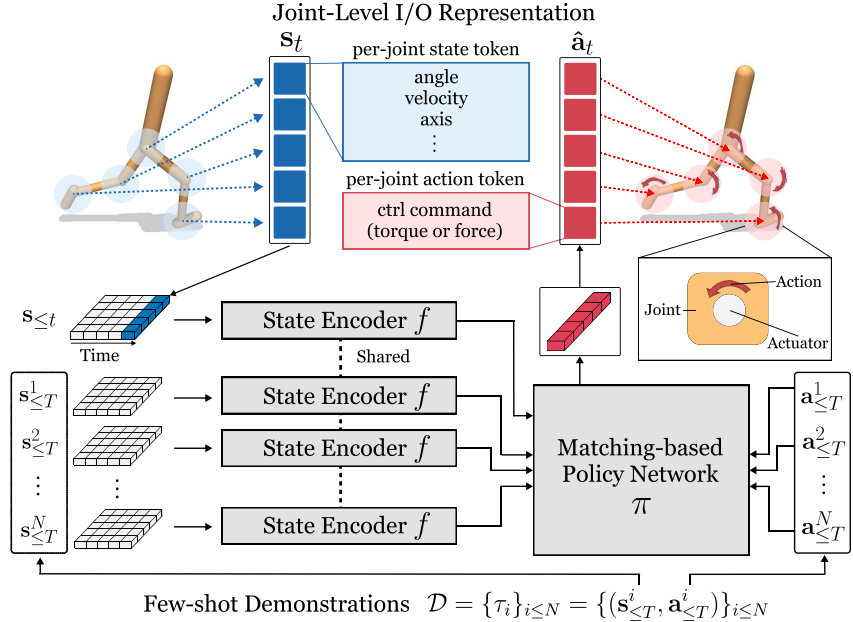}
    \caption{The overall framework of \modelname{}. First, the states and actions of various robot embodiments are tokenized into joint-level representations.
    The state tokens are then encoded by the state encoder to capture knowledge about the embodiments.
    Finally, a matching-based policy network uses few-shot demonstrations with the encoded state features to predict per-joint actions.}
    \label{fig:overall_framework}
\end{figure}

In this section, we introduce \modelname{}, a few-shot behavior cloning framework for simultaneous generalization of embodiments and tasks of continuous control.
Figure~\ref{fig:overall_framework} illustrates the overall framework.
\modelname{} addresses heterogeneous embodiments by tokenizing the states and actions into joint-level I/O (Section~\ref{sec:joint_io}) and employing a state encoder that captures knowledge about the structure and dynamics of the embodiment (Section~\ref{sec:state_encoder}).
Then, a matching-based policy network (Section~\ref{sec:policy_network}) predicts the action by leveraging a few given demonstrations.
The training protocol of \modelname{} consists of episodic meta-learning and few-shot fine-tuning (Section~\ref{sec:training}).


\subsection{Joint-Level I/O Representation}
\label{sec:joint_io}

To unify the state and action spaces of different embodiments, we adopt joint-level tokenization.
Joints are fundamental components of robots, and their primary source of action is the torque or force generated by actuators attached to each joint.
This allows us to standardize the states and actions of a robotic agent into per-joint observations and control commands.
Consequently, joint-level states and actions provide a natural modular representation, facilitating the compositional generalization of various robot embodiments\footnote{We consider a pre-defined set of joints whose \emph{compositions} differ per embodiment.}.

For any given embodiment $\mathcal{E}$, we represent the corresponding states $\mathbf{s}_t$ and actions $\mathbf{a}_t$ as arrays of joint-level tokens as follows:
\begin{equation}
    \mathbf{s}_t = \left[\mathbf{s}_{j, t}\right]_{j \le J_\mathcal{E}} \in \mathbb{R}^{J_\mathcal{E} \times d}, \quad \mathbf{a}_t = \left[\mathbf{a}_{j, t}\right]_{j \le J_\mathcal{E}} \in \mathbb{R}^{J_\mathcal{E} \times 1},
\end{equation}
where $J_\mathcal{E}$ is the number of joints in $\mathcal{E}$ and $t$ indicates the time.
The state token $\mathbf{s}_{j, t} \in \mathbb{R}^d$ represents per-joint information, such as position, velocity, movement axis, and motion types (\emph{i.e.,} linear or angular).
The action token $\mathbf{a}_{j, t} \in [-1, 1]$ represents the control command of $j$-th joint, where we assign zero value for free joints.
This joint-level representation ensures consistency across heterogeneous robotic embodiments, enabling unified learning of different continuous control problems.

\subsection{ \modelname{}}
\label{sec:ours}
Given the tokenized representation of joint states and actions, we aim to build an adaptive controller for arbitrary robot embodiments to perform arbitrary continuous control tasks based on a set of few demonstrations $\mathcal{D} = \{\tau^i\}_{i \le N}$.
To address the challenge discussed in Section~\ref{sec:challenges_and_desiderata}, we employ an encoder $f$ to extract the embodiment-aware state features and an adaptive policy network $\pi$ that efficiently leverages the demonstrations $\mathcal{D}$ to predict the actions.
\subsubsection{State Encoder for Embodiment Generalization}
\label{sec:state_encoder}

\begin{figure}[t!]
    \centering
    \includegraphics[width=0.8\linewidth]{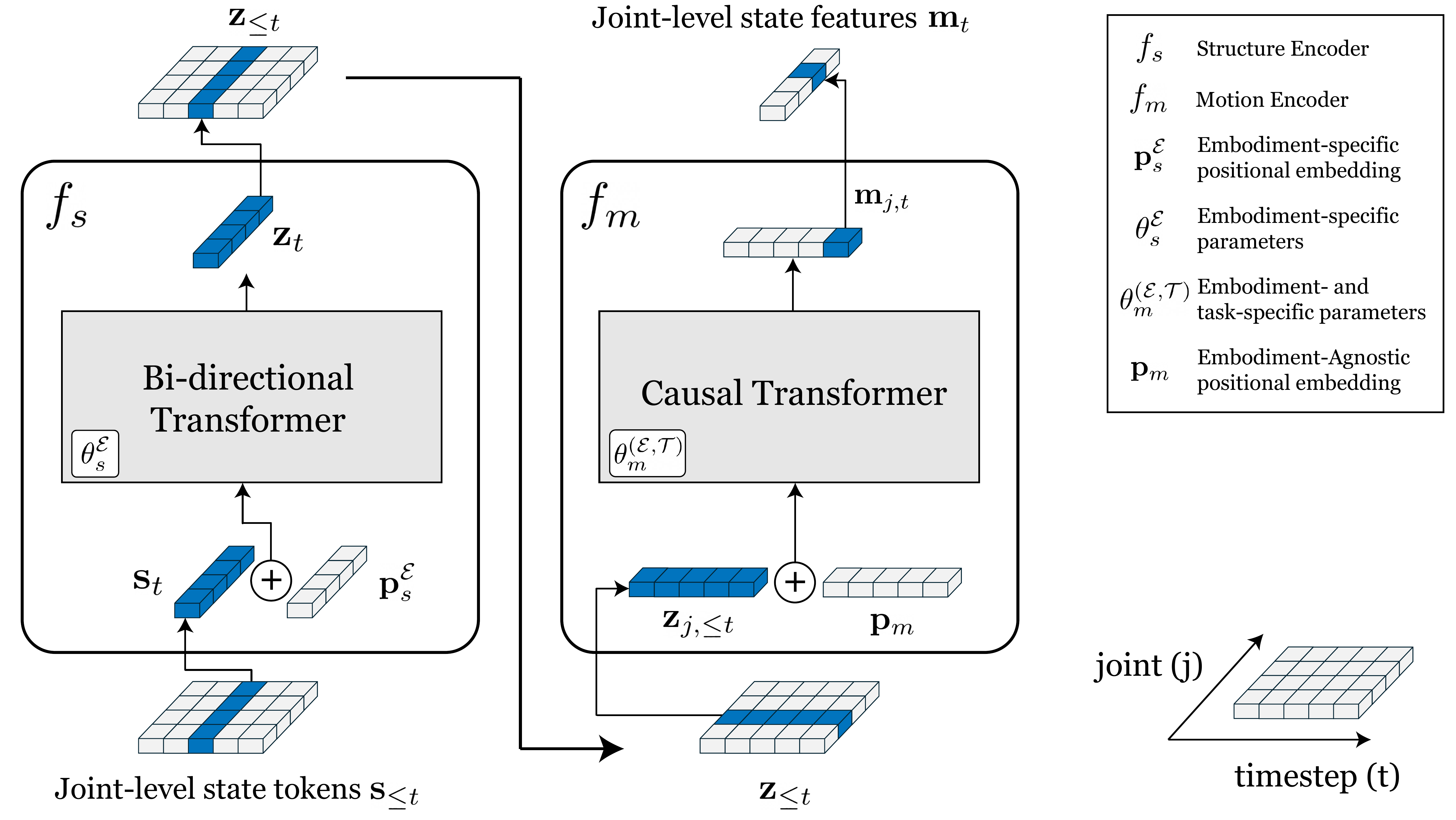}
    \caption{The state encoder $f$ consists of two component transformers.
    Joint-level state tokens are first encoded by the structure encoder $f_s$ along the joint axis, where the positional embedding and a part of backbone parameters adapt the model to each embodiment.
    The features are then passed to the motion encoder $f_m$, which computes causal attentions of per-joint features along the temporal axis, where a part of backbone parameters adapt the model both to the embodiment and task.}
    \label{fig:state_encoder}
\end{figure}

The state encoder $f$ encodes the tokens of the current state with history $\mathbf{s}_{\le t}$ into state features $\mathbf{m}_{t}$.
\begin{equation}
    \mathbf{m}_{t} = f(\mathbf{s}_{\le t}; \theta).
    \label{eqn:state_encoder}
\end{equation}
To effectively encode the states of each embodiment, we decompose our state encoder into two components: a structure encoder $f_s$ that captures morphological knowledge, and a motion encoder $f_m$ that captures dynamics knowledge.

\paragraph{Structure Encoder}
The structure encoder models the relationships among the joints within each embodiment.
As shown in Figure~\ref{fig:state_encoder}, we use a bi-directional transformer on the joint-level state tokens $\mathbf{s}_t$ at each timestep $t$ to extract the structure features $\mathbf{z}_t$:
\begin{align}
    \mathbf{z}_t &= f_s(\mathbf{s}_{t} + \mathbf{p}_s^\mathcal{E}; \theta_{s}, \theta_s^\mathcal{E}),
    \label{eqn:structure_encoder}
\end{align}
where the embodiment-specific positional embedding $\mathbf{p}_s^\mathcal{E}$ is added to the input tokens.
Note that we decompose the parameters of $f_s$ into adaptive parameters $\left(\mathbf{p}_s^\mathcal{E}, \theta_s^\mathcal{E}\right)$ that capture embodiment-specific knowledge and shared parameters $\theta_s$ that captures common knowledge across embodiments.
The positional embedding $\mathbf{p}_s^\mathcal{E}$ is crucial for adapting to local configurations (\emph{e.g.,} length, movement range) of joints in $\mathcal{E}$ not explicitly given in the state $\mathbf{s}_t$. 
Global configurations (\emph{e.g.,} control timestep) shared by all joints are handled through the embodiment-specific parameters $\theta_s^\mathcal{E}$ in the transformer backbone.
Shared parameters $\theta_s$ capture common knowledge, such as the physics governing the environment.

To enable efficient yet flexible adaptation during few-shot behavior cloning, we designate only a small portion of the backbone parameters to be embodiment-specific.
Inspired by parameter-efficient fine-tuning (PEFT) approaches that effectively modulate transformers with only a few parameters, we employ bias parameters~\cite{ben2022bitfit}, low-rank projection matrices~\cite{hu2022lora}, and also layer-scale parameters for $\theta_s^\mathcal{E}$.
As explained in Section~\ref{sec:training}, only the adaptive parameters $\left(\mathbf{p}_s^\mathcal{E}, \theta_s^\mathcal{E}\right)$ are updated during few-shot learning on unseen embodiments, ensuring robustness to overfitting the few demonstrations.



\paragraph{Motion Encoder}
While the state encoder $f_s$ encodes structural information about the embodiments, it does not model the temporal dynamics of states which is crucial for understanding continuous control tasks.
Therefore, we introduce a motion encoder $f_m$, which is a causal transformer that encodes the state features along the temporal axis.
As illustrated in Figure~\ref{fig:state_encoder}, $f_m$ rearranges the encoded structure features $\mathbf{z}_{\le t}$ into separate temporal sequences of joint-level features $\mathbf{z}_{j, \le t}$, then produce the motion features for each joint $\mathbf{m}_{j, t}$ auto-regressively:
\begin{align}
    \mathbf{m}_{j, t} &= f_m(\mathbf{z}_{j, \le t} + \mathbf{p}_m; \theta_m, \theta_m^{(\mathcal{E}, \mathcal{T})}), \quad \forall j \le J_\mathcal{E}
    \label{eqn:motion_encoder}
\end{align}
where $\mathbf{p}_m$ denotes the positional embedding for specifying the timesteps $t$.
Additionally, we introduce a small portion of adaptive parameters $\theta_m^{(\mathcal{E}, \mathcal{T})}$ in the causal transformer backbone, which is both specific to embodiment $\mathcal{E}$ and task $\mathcal{T}$.
These parameters help the model understand the unique motions in the few-shot demonstrations that are specific to each task and embodiment.
We use the same PEFT techniques employed in the structure encoder for the adaptive parameters.

\subsubsection{Few-shot Policy Adaptation for Task Generalization}
\label{sec:policy_network}

\begin{figure}[t!]
    \centering
    \includegraphics[width=\linewidth]{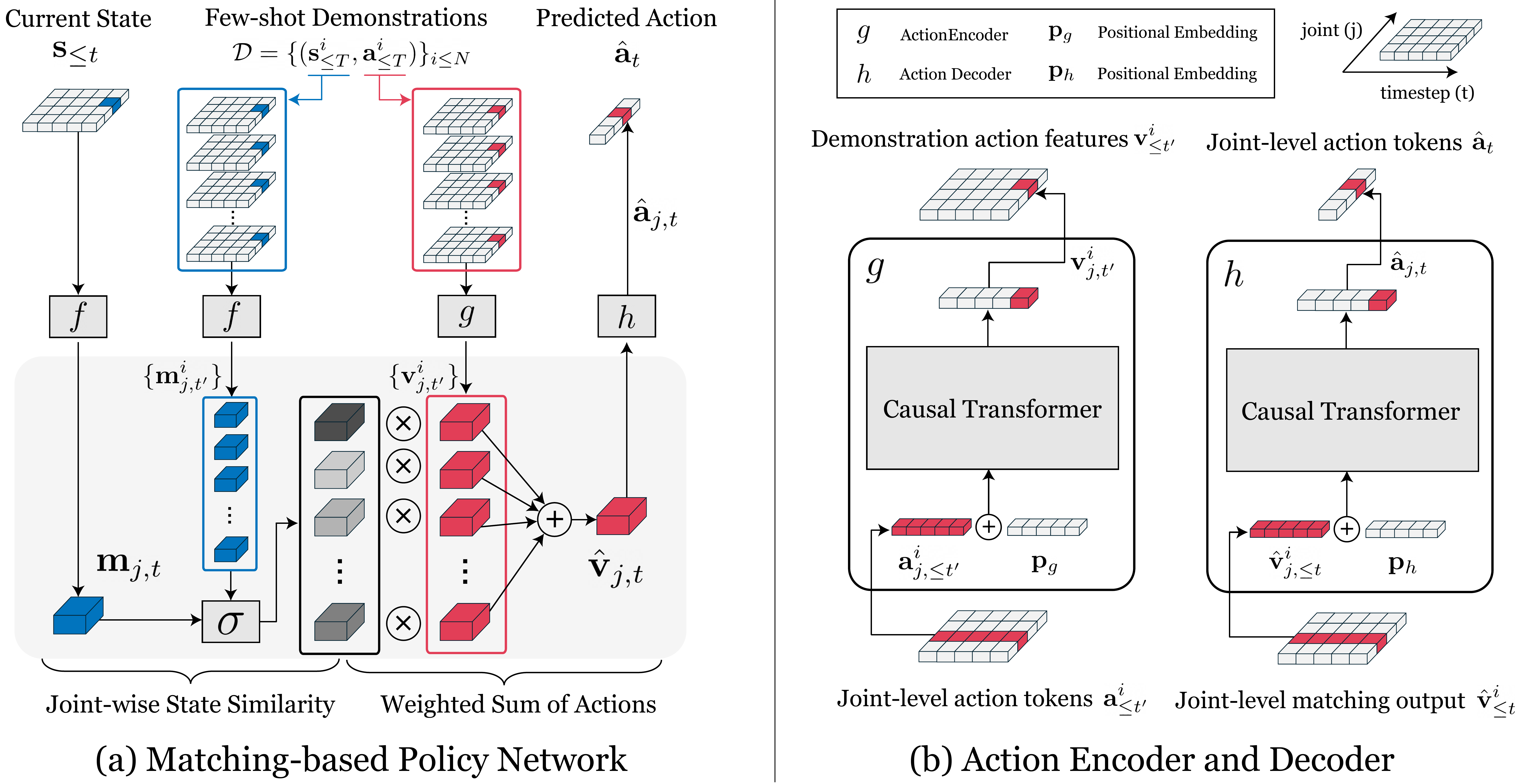}
    \caption{An illustration of the matching-based policy network $\pi$.
    (a) Each state and action token in few-shot demonstrations is encoded by the corresponding encoders $f$ and $g$, where we use the same encoder $f$ used for the current state.
    A matching module $\sigma$ then computes the weighted sum of action features based on the joint-wise similarity between state features.
    Finally, an action decoder $h$ decodes the joint-wise matching output to predict the current action.
    (b) Both the action encoder $g$ and decoder $h$ are causal transformers operating along the temporal axis of action tokens and features.}
    \label{fig:policy_network}
\end{figure}

To learn unseen tasks $\mathcal{T}$, we design an adaptive policy network $\pi$ with a matching framework~\cite{kim2022universal,kim2024chameleon} that incorporates the demonstrations $\mathcal{D} = \{(\mathbf{s}_{\le T}^i, \mathbf{a}_{\le T}^i)\}_{i \le N}$ to produce an action $\hat{\mathbf{a}}_t$.
\begin{equation}
    \hat{\mathbf{a}}_t = \pi(\mathbf{m}_{\le t}, \mathcal{D}; \phi).
    \label{eqn:policy_network}
\end{equation}
Figure~\ref{fig:policy_network} illustrates the architecture of the policy network, which consists of an action encoder $g$, a matching module $\sigma$, and an action decoder $h$.

\paragraph{Matching-based Policy Network}
To incorporate the demonstrations, we encode the state and action tokens in the demonstrations using the state encoder $f$ introduced in Section~\ref{sec:state_encoder} and an additional action encoder $g$, respectively.
\begin{align}
    \mathbf{m}_{j, t^\prime}^i &= f(\mathbf{s}_{j, \le t^\prime}^i; \theta), \quad\qquad\forall j \le J_\mathcal{E}, ~\forall t^\prime \le T, ~\forall i \le N,
    \label{eqn:demonstration_encoding_state}\\
    \mathbf{v}_{j, t^\prime}^i &= g(\mathbf{a}_{j, \le t^\prime}^i + \mathbf{p}_g; \phi_g), \quad\forall j \le J_\mathcal{E}, ~\forall t^\prime \le T, ~\forall i \le N,
    \label{eqn:demonstration_encoding_action}
\end{align}
where we employ a causal transformer for $g$ with temporal position embedding $\mathbf{p}_g$.
Then $\pi$ incorporates the motion features $\mathbf{m}_t$ of the current state (Eq.~\eqref{eqn:state_encoder}) and the encoded demonstration features (Eq.~\eqref{eqn:demonstration_encoding_state} and \eqref{eqn:demonstration_encoding_action}) via joint-wise matching as follows:
\begin{equation}
    \hat{\mathbf{v}}_{j, t} = \sum_{t^\prime \le T} \sum_{i \le N} \sigma(\mathbf{m}_{j, t}, \mathbf{m}_{j, t^\prime}^i) \cdot \mathbf{v}_{j, t^\prime}^i, \quad \forall j \le J_\mathcal{E},
    \label{eqn:matching}
\end{equation}
where $\sigma$ is a similarity function (\emph{e.g.}, cosine similarity).
Finally, we employ an action decoder $h$, a causal transformer that decodes the joint-wise matching output $\hat{\mathbf{v}}_{j, t}$ into $j$-th action token $\hat{\mathbf{a}}_{j, t}$.
\begin{equation}
    \hat{\mathbf{a}}_{j, t} = h(\hat{\mathbf{v}}_{j, \le t}; \phi_h),  \quad \forall j \le J_\mathcal{E}.
\end{equation}
The matching-based policy network offers significant benefits for few-shot behavior cloning, particularly when dealing with unseen tasks and embodiments.
Its robust adaptation capabilities stem from effectively incorporating demonstration data through a similarity function, which dynamically matches current state features with those from demonstrations.
This non-parametric approach minimizes overfitting, enhancing generalization from limited examples.

\paragraph{Hierarchical IL Interpretation}
Eq.~\eqref{eqn:matching} can also be interpreted as hierarchical imitation learning that generalizes to unseen tasks using a transferable skill set.
Since the action encoder $g$ extracts a pool of temporal action features that are composed to produce the current action $\mathbf{a}_t$ in the feature space, we can treat the action features $\mathbf{v}_{j, t^\prime}^i$ as \emph{local motor skills}, \emph{i.e.,} building blocks of joint behavior for various control tasks.
This modular approach lets the network recombine these skills to efficiently tackle new challenges.
By assigning high similarity scores to relevant demonstrations, the policy network ensures accurate imitation of expert behavior, improving performance on novel tasks.

\subsection{Training \& Inference}
\label{sec:training}

The training protocol of \modelname{} consists of two stages: episodic meta-learning and few-shot fine-tuning, where we train whole parameters $(\theta, \phi)$ of the model during the first stage while we train only the adaptive parameters $(\mathbf{p}_s^\mathcal{E}, \theta_s^\mathcal{E}, \theta_m^{(\mathcal{E}, \mathcal{T})})$ during the second stage.

\paragraph{Episodic Meta-Learning}
During episodic meta-learning, the model acquires general knowledge of continuous control through a number of episodes that mimic few-shot behavior cloning scenarios to ensure effective adaptation to unseen embodiments and tasks.
To this end, we leverage a meta-training dataset that consists of demonstrations $\mathcal{B}_{(\mathcal{E}, \mathcal{T})}$ by expert agents with various embodiments $\mathcal{E}$ and tasks $\mathcal{T}$.
At each episode, we first sample a continuous control problem $(\mathcal{E}, \mathcal{T})$, then sample two subsets of $\mathcal{B}_{(\mathcal{E}, \mathcal{T})}$: a set of support data $\mathcal{D}_S$ and query data $\mathcal{D}_Q$.
Then the model is trained to imitate query data using the demonstrations.
\begin{equation}
    \min_{(\theta, \phi)}~ \mathbb{E}_{p(\mathcal{E}, \mathcal{T})} \Bigg[ \mathbb{E}_{\mathcal{D}_S, \mathcal{D}_Q \sim \mathcal{B}_{(\mathcal{E}, \mathcal{T})}} \Bigg[
    \frac{1}{|\mathcal{D}_Q|} \sum_{(\mathbf{s}_t, \mathbf{a}_t) \in \mathcal{D}_Q} \left\| \mathbf{a}_t - \hat{\mathbf{a}}_t \right\|^2
    \Bigg] \Bigg],
    \label{eqn:episodic_meta_learning}
\end{equation}
where $\hat{\mathbf{a}}_t$ is produced by Eq.~\eqref{eqn:policy_network} using $\mathcal{D}_S$, and $p(\mathcal{E}, \mathcal{T})$ is a uniform distribution over all control problems within the dataset.

\paragraph{Few-shot Fine-Tuning}
After acquiring the meta-knowledge about continuous control problems, we apply our \modelname{} in a few-shot behavior cloning setup, where it should adapt to both unseen embodiments and tasks with a few demonstrations $\mathcal{D}$.
To this end, we randomly split $\mathcal{D}$ into two disjoint subsets, and fine-tune the model with Eq.~\eqref{eqn:episodic_meta_learning} but \emph{with only respect to} the embodiment-specific and task-specific parameters $(\mathbf{p}_s^\mathcal{E}, \theta_s^\mathcal{E}, \theta_m^{(\mathcal{E}, \mathcal{T})})$ while freezing the rest.
After fine-tuning, the model uses the whole demonstrations $\mathcal{D}$ in the policy network to produce the actions at evaluation.





\section{Related Work}
\label{sec:related_work}

\paragraph{Modular Policy Learning}
Modular policy learning aims to develop modular policies for multi-joint robots that are adaptable to various morphologies.
NerveNet~\cite{wang2018nervenet} uses a Graph Neural Network (GNN) to model structural relationships between joint-level features.
SWAT~\cite{hong2021structure} leverages graph features like the normalized graph Laplacian for improved structural learning in reinforcement learning (RL).
Amorpheous~\cite{kurin2021my} and MetaMorph~\cite{gupta2021metamorph} use transformer architectures, treating the morphological graph as fully connected to exploit transformers' capabilities.
These works focus on RL and zero-shot generalization to specific locomotion tasks. Furuta~et~al.~\cite{furuta2022system} proposes an imitation learning framework with a benchmark environment for extensive morphologies, showing promising zero-shot generalization but limited to specific tasks like reaching a goal position, and not handling general continuous control tasks.


\paragraph{Few-shot Imitation Learning}
Few-shot Imitation Learning (IL) approaches focus on generalizing novel RL tasks with only a few demonstrations.
Gradient-based meta-learning algorithms like MAML~\cite{finn2017model} and Bayesian MAML~\cite{yoon2018bayesian} have been explored for rapid task adaptation.
Duan~et~al.~\cite{duan2017one} addressed one-shot imitation learning by incorporating an attention model over the query state and demonstrations.
FIST~\cite{hakhamaneshi2021hierarchical} is a hierarchical skill transition model that learns to extract transferable high-level skill sets from demonstrations and leverages this skill knowledge during adaptation.
Recently, PDT~\cite{xu2022prompting} introduced a transformer-based few-shot learner, using the few-shot demonstration as a prompt token.
Similarly, HDT~\cite{xu2022hyper} trains a hyper-network to generate adapter parameters for a pre-trained decision transformer, adapting to few-shot demonstrations.
Despite notable generalization abilities, these few-shot IL approaches have only been studied within a single embodiment, limiting their applicability to real-world scenarios with heterogeneous embodiments.
\section{Experiments}
\label{sec:experiments}
\subsection{Experimental Setup}

\paragraph{Environment and Dataset}
We evaluate the few-shot behavior cloning of unseen embodiments and tasks within the DeepMind Control (DMC) suite~\cite{tunyasuvunakool2020dmc}, which includes continuous control tasks featuring diverse kinematic structures.
We select 30 tasks from 10 embodiments as training tasks and 8 tasks from 4 embodiments as held-out evaluation tasks.
The evaluation tasks include three unseen embodiments (\texttt{hopper}, \texttt{reacher-four}, \texttt{wolf}) and one seen embodiment (\texttt{walker}), with the \texttt{wolf} being an additional embodiment introduced in \cite{wang2018nervenet}.
Our meta-training dataset is constructed using a replay buffer of an expert agent~\cite{yarats2021image}, consisting of up to 2000 demonstration trajectories for each task and embodiment.
Each demonstration consists of state-action pairs over $T=500$ timesteps, with rewards discarded for both episodic meta-learning and few-shot behavior cloning.
For $N$-shot few-shot behavior cloning, we use the last $N$ demonstrations from the buffer.
A full detail of the dataset is included in Appendix~\ref{appendix:data_details}.

\paragraph{Evaluation Protocol}
We evaluate all models with 20 different initial states and report the mean and standard error.
For evaluation metric, we use a \emph{normalized score}~\cite{fu2020d4rl} calculated by $\frac{\text{score} - \text{random score}}{\text{expert score} - \text{random score}}$, where each score represents the average cumulative rewards during the evaluation.
The random score is obtained by evaluating a random agent using a uniform distribution policy over the action spaces.
We evaluate the models every 1000 iterations of few-shot behavior cloning and report the best score.
For models that use task-specific low-rank projection matrices, we search the rank parameter over $\{4, 8, 16\}$ and report the best one.
Throughout the experiments, we present 5-shot learning results ($N=5$) unless otherwise specified.

\paragraph{Baselines}
We compare our model with various Decision Transformer~\cite{chen2021decision} (DT)-based few-shot imitation learning approaches and two transformer-based modular policy learning approaches.
In the DT-based models, we exclude return-to-go tokens from the input tokens to simulate behavior cloning.
Since the DT architecture is not inherently designed to handle heterogeneous state and action spaces, we modify its input and output linear heads in an embodiment-specific manner and fine-tune them for unseen embodiments.

\begin{itemize}[leftmargin=*]
\item \textbf{DT-based Models.}
From-Scratch Decision Transformer \textbf{(FS-DT)} is a decision transformer that trains a downstream task directly from randomly initialized weights.
\textbf{Multi-Task Decision Transformer (MT-DT)} is a variant of the decision transformer that trains multiple tasks with task-specific parameters and fine-tunes only these parameters for few-shot adaptation.
For the task-specific parameters, we use the same parameters as $\theta_m^{(\mathcal{E}, \mathcal{T})}$ used in our motion encoder.
Prompt-based Decision Transformer \textbf{(PDT)}~\cite{xu2022prompting} adapts its policy by conditioning on the few-shot demonstrations through prompting.
We also report the performance of a variant of PDT that fine-tunes task-specific parameters (\textbf{PDT+PEFT}), similar to MT-DT.
Hyper Decision Transformer \textbf{(HDT)}~\cite{xu2022hyper} employs a hyper-network conditioned on few-shot demonstrations to generate parameters of the Adapter~\cite{mahabadi2021parameter} module applied to DT.
Learning To Modulate \textbf{(L2M)}~\cite{schmied2024learning} incorporates parameter-efficient fine-tuning (PEFT) techniques to DT architecture.
While L2M is not directly proposed for few-shot imitation learning, we include this baseline since it uses a similar PEFT technique as ours.
\item \textbf{Modular Policy-based Models.} We include two modular policy learning approaches, \textbf{MetaMorph}~\cite{gupta2021metamorph} and \textbf{MTGv2}~\cite{furuta2022system},
which utilize a transformer architecture to encode joint-level states.
Originally designed for zero-shot learning of locomotion tasks, we adapt these approaches by incorporating task-specific linear heads and fine-tuning them on few-shot demonstrations.
Both models are trained using the behavior cloning (BC) objective as described in \cite{furuta2022system}.
\end{itemize}

\paragraph{Implementation Details}
We implement both the structure encoder and the motion encoder using a 6-layer transformer with 4 attention heads and a hidden dimension of 512.
Due to the quadratic computation cost of the transformer, we set the maximum history size of causal attention layers in the encoders to 10.
The baseline models use the same transformer backbone.
All models are trained for 200,000 iterations on the meta-training dataset and fine-tuned for 10,000 iterations on the few-shot demonstrations.
For downstream embodiments that are structurally similar to a training task (\emph{e.g.}, \texttt{reacher-three} and \texttt{reacher-four}), we initialize the encoder's embodiment-specific parameters using the trained parameters during fine-tuning.
More details are included in Appendix~\ref{appendix:implementation_details}.

\subsection{Main Results}

\begin{table*}[!t]
\caption{$5$-shot behavior cloning results on DeepMind Control (DMC) suite.}
\label{tab:main_table}
\begin{center}
    \renewcommand{\arraystretch}{1.2}
    \renewcommand{\aboverulesep}{0pt}
    \renewcommand{\belowrulesep}{0pt}
    \setlength\tabcolsep{4pt}
    \scriptsize
    \begin{tabular}{Rc@{\hskip 0.2cm}c@{\hskip 0.2cm}c|c@{\hskip 0.2cm}c|c@{\hskip 0.2cm}RRc}
        \toprule
        \multirow{1}{*}{} &
        \multicolumn{7}{R}{Unseen Emb.} &
        Seen Emb. & 
        \multirow{3}{*}{Avg.} \\
        \cmidrule{1-9}
        Emb. ($\mathcal{E}$) &
        \multicolumn{3}{c|}{\texttt{hopper}} & 
        \multicolumn{2}{c|}{\texttt{wolf}} & 
        \multicolumn{2}{R}{\texttt{reacher-four}} & 
        \multicolumn{1}{R}{\texttt{walker}} &
        \\

        \cmidrule{1-9}
        
        Task ($\mathcal{T}$) &
        \texttt{hop} &
        \texttt{hop-bwd.} &
        \texttt{stand} &
        \texttt{walk} &
        \texttt{run} &
        \texttt{easy} & 
        \texttt{hard} & 
        \texttt{walk-bwd.} &
        \\
        \midrule
        FS-DT~\cite{chen2021decision} &
        1.6$\pm$1.5 &
        56.9$\pm$6.2 &
        9.8$\pm$0.9 &
        44.7$\pm$10.8 &
        35.8$\pm$7.2 &
        -0.7$\pm$4.2 &
        5.5$\pm$4.9 &
        20.6$\pm$8.6 &
        21.8 \\
        
        MT-DT &
        3.2$\pm$2.3 &
        64.3$\pm$7.0 &
        10.1$\pm$1.1 &
        53.4$\pm$11.8 &
        46.2$\pm$10.2 &
        10.2$\pm$7.4 &
        8.4$\pm$5.1 &
        \textbf{86.4$\pm$5.0} &
        35.3 \\
        
        PDT~\cite{xu2022prompting} &
        0.9$\pm$0.7 &
        29.7$\pm$9.3 &
        6.5$\pm$1.9 &
        47.0$\pm$11.4 &
        35.7$\pm$8.3 &
        3.3$\pm$4.7 &
        5.8$\pm$3.5 &
        5.0$\pm$1.5 &
        16.7 \\
        
        PDT+PEFT &
        2.3$\pm$1.5 &
        61.7$\pm$7.7 &
        12.3$\pm$4.0 &
        54.9$\pm$8.4 &
        52.7$\pm$6.4 &
        -1.7$\pm$3.3 &
        9.1$\pm$5.9 &
        74.1$\pm$8.5 &
        33.2 \\
        
        HDT~\cite{xu2022hyper} &
        0.8$\pm$0.4 &
        51.6$\pm$7.9 &
        13.7$\pm$3.7 &
        56.4$\pm$11.0 &
        38.7$\pm$9.1 &
        1.2$\pm$5.3 &
        8.5$\pm$5.5 &
        3.3$\pm$1.0 &
        21.8 \\
        
        L2M~\cite{schmied2024learning} &
        4.5$\pm$1.7 &
        45.1$\pm$7.9 &
        4.0$\pm$1.2 &
        50.6$\pm$10.7 &
        65.9$\pm$3.8 &
        3.7$\pm$6.3 &
        15.5$\pm$7.4 &
        40.0$\pm$9.8 &
        28.7 \\
        
        \midrule
        MetaMorph~\cite{gupta2021metamorph}&
        33.4$\pm$4.9 &
        81.8$\pm$2.4 &
        54.1$\pm$4.7 &
        73.0$\pm$4.4 &
        29.7$\pm$4.6 &
        -4.1$\pm$2.6 &
        3.8$\pm$3.1 &
        31.8$\pm$8.6 &
        37.9 \\
        
        MTGv2~\cite{furuta2022system} &
        21.6$\pm$3.3 &
        83.1$\pm$2.2 &
        40.7$\pm$6.3 &
        68.2$\pm$7.3 &
        29.1$\pm$4.3 &
        -1.4$\pm$2.7 &
        4.6$\pm$4.8 &
        35.5$\pm$5.9 &
        35.2 \\
        \midrule
        Ours &
        \textbf{49.1$\pm$6.1} &
        \textbf{87.2$\pm$1.6} &
        \textbf{82.5$\pm$4.9} &
        \textbf{91.7$\pm$5.1} &
        \textbf{67.3$\pm$3.1} &
        \textbf{56.1$\pm$8.8} &
        \textbf{50.8$\pm$10.6} &
        84.3$\pm$5.7 &
        \textbf{71.1} \\

        \bottomrule
    \end{tabular}
\end{center}
\end{table*}

Table~\ref{tab:main_table} shows the $5$-shot behavior cloning results of the models on both unseen and seen embodiments.
We observe that \modelname{} consistently outperforms all baselines across various continuous control problems, demonstrating its effectiveness.
Existing few-shot imitation learning approaches that lack an embodiment generalization mechanism, such as PDT and HDT, struggle to adapt to unseen embodiments, often performing comparably to the naive from-scratch baseline (DT-FS).
In contrast, modular policy learning approaches like MetaMorph and MTGv2 show better adaptation to unseen embodiments (\emph{e.g.}, \texttt{hopper}).
However, their performance is still inferior to \modelname{},  which can be attributed to the absence of a few-shot adaptation mechanism for unseen tasks.

\begin{figure}[t!]
    \centering
    \includegraphics[width=\linewidth]{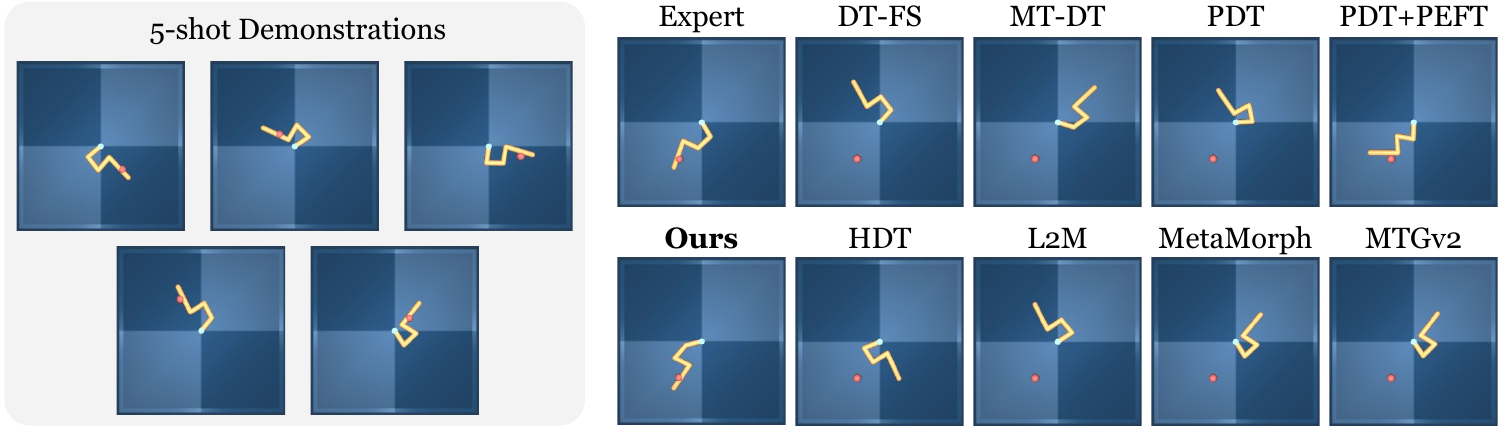}
    \caption{Qualitative comparison on the \texttt{hard} task of the \texttt{reacher-four} embodiment, visualizing the final states of the demonstrations and the rollout trajectories of each model.
    In this task, the robot must move its limb tip to the goal position (visualized as a red ball).
    While most of the baselines converge to one of the poses in the demonstrations and ignore the goal position, our model accurately solves the task with a distinct pose from the demonstrations.
    }
    \label{fig:reacher_qualitative}
\end{figure}

Notably, our model significantly outperforms all baselines in the challenging \texttt{reacher-four} embodiment, where models must understand the notion of a \emph{goal position} given only five demonstrations.
Figure~\ref{fig:reacher_qualitative} shows that while baseline models converge to the pose in the demonstrations regardless of the goal position, our model successfully reaches the goal position and converges with a unique pose.
We attribute this success to the modular nature of our model.
Our state encoder effectively shares knowledge about joint-level motions thanks to its parametrization, and the matching-based policy network flexibly exploits the local motor skills from a few demonstrations.
We provide more results and analysis in Appendix~\ref{appendix:additional_results}.

\subsection{Ablation Studies}
\label{sec:ablation_study}


\begin{figure}[t!]
    \centering
    \includegraphics[width=\linewidth]{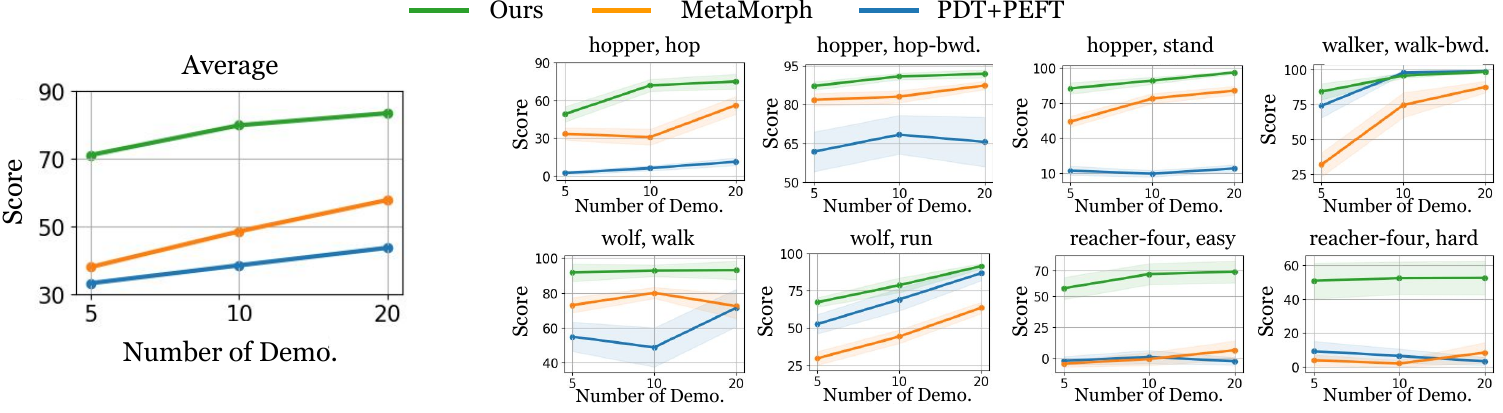}
    \caption{Ablation study on the number of demonstrations.
    We plot the normalized scores for each pair of embodiment and task $(\mathcal{E}, \mathcal{T})$ and their average, varying the number of shots as 5, 10, and 20.}
    \label{fig:shot_ablation}
\end{figure}

\paragraph{Ablation on the Architecture}
To verify the effectiveness of each architectural component introduced in Section~\ref{sec:method}, we conduct an ablation study by progressively replacing each module—the structure encoder $f_s$, the motion encoder $f_m$, and the matching module $\sigma$—with a linear layer.
To isolate the effect of adaptation parameters, we replace $f_s$ with task-specific linear layers and $f_m$ with embodiment- and task-specific linear layers.
Table~\ref{tab:model_component_ablation} summarizes the results.
We observe that the structure encoder $f_s$ is crucial for generalization to unseen embodiments, as performance drops drastically when it is removed (row 1 vs. row 3).
This indicates that the structure encoder captures modular knowledge about various morphologies, transferrable to unseen embodiments.
Combined with the structure encoder, the motion encoder further improves performance (row 2 vs. row 3), especially on the seen embodiment (\texttt{walker}).
This shows that modeling the temporal relationships of joints is beneficial when the model understands the embodiment.
Finally, the matching module consistently improves performance (row 3 vs. row 4), particularly on challenging tasks such as the \texttt{hop} task of \texttt{hopper} and the tasks of \texttt{reacher-four}
This demonstrates the effectiveness of the matching architecture in preventing overfitting with few demonstrations.
We provide more ablation studies on architectural components in Appendix~\ref{appendix:more_ablation_architecture}.

\paragraph{Ablation on the Parametrization}
To analyze the impact of the embodiment-specific and task-specific parametrization introduced in the state encoder, we conduct an ablation study by removing the adaptive parameters $\mathbf{p}_s^\mathcal{E}, \theta_s^\mathcal{E}$ in the structure encoder and $\theta_m^{(\mathcal{E}, \mathcal{T})}$ the motion encoder.
In this study, the matching module $\sigma$ is used in all models, where we provide additional results without using it in Appendix~\ref{appendix:more_ablation_adaptation}.
The results show that the model without adaptive parameters in the structure encoder (row 1) performs well in many tasks, likely due to the universality of joint-level input/output representation, which allows for compositional generalization.
However, adding embodiment-specific parameters consistently improves performance across all tasks, indicating the benefit of capturing embodiment-specific knowledge. 
The model without adaptive parameters in the motion encoder (row 2) remains competitive with the model including them (row 3) in many tasks but fails in the \texttt{hop} task of \texttt{hopper} and the \texttt{walk-bwd.} task of \texttt{walker}.
This failure, even in a seen embodiment (\texttt{walker}), suggests that certain unique motions cannot be adequately captured without adaptive parameters.
Overall, introducing adaptive parameters in both encoders yields the best performance.



\paragraph{Ablation on the Number of Demonstrations}
In Figure~\ref{fig:shot_ablation}, we plot the performance of our model and the two best-performing baselines—one from modular policy learning and the other from few-shot IL—by varying the number of demonstrations.
As expected, the performance of our model consistently increases with the number of demonstrations provided.
The consistent superiority of our model in few-shot behavior cloning, across varying numbers of demonstrations, highlights the necessity of (1) a powerful encoder capable of handling unseen embodiments, and (2) a few-shot policy adaptation mechanism that enables robust few-shot learning on unseen tasks.
These components are crucial for achieving simultaneous few-shot generalization to both unseen embodiments and tasks.

We provide more ablation studies on meta-training task composition in Appendix~\ref{appendix:more_ablation_task_composition} 

\begin{table*}[!t]
\caption{Ablation study on the architectural components. 
}
\label{tab:model_component_ablation}
\begin{center}
    \renewcommand{\arraystretch}{1.2}
    \renewcommand{\aboverulesep}{0pt}
    \renewcommand{\belowrulesep}{0pt}
    \setlength\tabcolsep{6pt}
    \scriptsize
    \begin{tabular}{ccRc@{\hskip 0.2cm}c@{\hskip 0.2cm}c|c@{\hskip 0.2cm}c|c@{\hskip 0.2cm}RRc}
        \toprule
        \multirow{3}{*}{$f_s$} &
        \multirow{3}{*}{$f_m$} &
        \multirow{3}{*}{$\sigma$} &
        \multicolumn{7}{R}{Unseen Emb.} & Seen Emb. & 
        \multirow{3}{*}{Avg.} \\
        \cmidrule{4-11}
        
        & & & \multicolumn{3}{c|}{\texttt{hopper}} & 
        \multicolumn{2}{c|}{\texttt{wolf}} & 
        \multicolumn{2}{R}{\texttt{reacher-four}} & 
        \multicolumn{1}{R}{\texttt{walker}} & \\
        \cmidrule{4-11}
        
        & & &
        \texttt{hop} &
        \texttt{hop-bwd.} &
        \texttt{stand} &
        \texttt{walk} &
        \texttt{run} &
        \texttt{easy} & 
        \texttt{hard} & 
        \texttt{walk-bwd.} &
        \\

        \midrule
        
         \xmark & \cmark &\xmark &
        5.3$\pm$3.8 &
        46.7$\pm$9.3 &
        14.9$\pm$5.9 &
        64.5$\pm$11.1 &
        53.5$\pm$8.0 &
        -3.2$\pm$3.6 &
        4.3$\pm$5.1 &
        66.6$\pm$10.5 &
        31.6 \\

        \cmark & \xmark & \xmark &
        13.1$\pm$5.6 &
        \textbf{88.1$\pm$1.7} &
        75.0$\pm$5.7 &
        71.8$\pm$6.9 &
        53.7$\pm$8.0 &
        4.1$\pm$7.2 &
        3.4$\pm$4.1 &
        48.1$\pm$5.4 &
        44.7 \\

        \cmark & \cmark & \xmark &
        22.8$\pm$5.9 &
        86.3$\pm$1.6 &
        \textbf{83.4}$\pm$4.2 &
        79.7$\pm$6.4 &
        57.3$\pm$6.7 &
        54.9$\pm$9.5 &
        24.8$\pm$8.2 &
        73.4$\pm$5.0 &
        60.3 \\
        \midrule
        
        \cmark & \cmark & \cmark &
        \textbf{49.1$\pm$6.1} &
        87.2$\pm$1.6 &
        82.5$\pm$4.9 &
        \textbf{91.7$\pm$5.1} &
        \textbf{67.3$\pm$3.1} &
        \textbf{56.1$\pm$8.8} &
        \textbf{50.8$\pm$10.6} &
        \textbf{84.3$\pm$5.7} &
        \textbf{71.1} \\

        \bottomrule
    \end{tabular}
\end{center}

\vspace{0.5cm}

\caption{Ablation study on the adaptive parameters in the state encoder.}
\label{tab:tsparam_ablation}
\begin{center}
    \renewcommand{\arraystretch}{1.2}
    \renewcommand{\aboverulesep}{0pt}
    \renewcommand{\belowrulesep}{0pt}
    \setlength\tabcolsep{6pt}
    \scriptsize
    \begin{tabular}{@{\hskip 0.1cm}c@{\hskip 0.2cm}Rc@{\hskip 0.2cm}c@{\hskip 0.2cm}c|c@{\hskip 0.2cm}c|c@{\hskip 0.2cm}RRc}
        \toprule
        \multirow{3}{*}{$\mathbf{p}_s^{\mathcal{E}}, \theta_s^\mathcal{E}$} &
        \multirow{3}{*}{$\theta_m^{(\mathcal{E}, \mathcal{T})}$} &
        \multicolumn{7}{R}{Unseen Emb.} &
        Seen Emb. &
        \multirow{3}{*}{Avg.} \\
        \cmidrule{3-10}
        
        & & \multicolumn{3}{c|}{\texttt{hopper}} & 
        \multicolumn{2}{c|}{\texttt{wolf}} & 
        \multicolumn{2}{R}{\texttt{reacher-four}} & 
        \multicolumn{1}{R}{\texttt{walker}} &
        \\

        \cmidrule{3-10}
        
        & & 
        \texttt{hop} &
        \texttt{hop-bwd.} &
        \texttt{stand} &
        \texttt{walk} &
        \texttt{run} &
        \texttt{easy} & 
        \texttt{hard} & 
        \texttt{walk-bwd.} &
        \\
        \midrule

        \xmark & \hspace{-0.3cm} \cmark \hspace{-0.3cm} &
        37.6$\pm$5.6 &
        76.8$\pm$3.6 &
        68.0$\pm$5.6 &
        85.4$\pm$4.5 &
        47.2$\pm$2.9 &
        17.4$\pm$8.9 &
        0.4$\pm$0.9 &
        59.2$\pm$8.1 &
        49.0 \\
        
        \cmark & \hspace{-0.3cm} \xmark \hspace{-0.3cm} &
        16.4$\pm$6.4 &
        84.7$\pm$2.7 &
        \textbf{84.0$\pm$3.7} &
        89.2$\pm$2.6 &
        \textbf{71.0$\pm$3.1} &
        \textbf{70.8$\pm$7.9} &
        48.1$\pm$9.4 &
        5.3$\pm$1.1 &
        58.7 \\
        \midrule
        
        \cmark & \hspace{-0.3cm} \cmark \hspace{-0.3cm} &
        \textbf{49.1$\pm$6.1} &
        \textbf{87.2$\pm$1.6} &
        82.5$\pm$4.9 &
        \textbf{91.7$\pm$5.1} &
        67.3$\pm$3.1 &
        56.1$\pm$8.8 &
        \textbf{50.8$\pm$10.6} &
        \textbf{84.3$\pm$5.7} &
        \textbf{71.1} \\
        \bottomrule
    \end{tabular}

\end{center}
\end{table*}

\section{Conclusion}
\label{sec:conclusion}

We addressed the challenging problem of few-shot behavior cloning with unseen embodiments and tasks in continuous control.
Our framework, \modelname{}, effectively handles diverse embodiments using two key components: the state encoder and the matching-based policy network.
Leveraging the modular nature of joint-level input/output representations, our state encoder extracts transferable features about the morphology and dynamics of the embodiment, capturing both specific and shared knowledge.
The matching-based policy network uses these features to infer task structures from few-shot demonstrations, enabling robust imitation without overfitting.
Experiments showed that our model generalizes well to unseen embodiments and tasks with only five demonstrations.

\begin{ack}
This work was in part supported by the National Research Foundation of Korea (RS-2024-00351212 and RS-2024-00436165), the Institute of Information \& communications Technology Planning \& Evaluation (IITP) (RS-2022-II220926, RS-2024-00509279, RS-2021-II212068, RS-2022-II220959, and RS-2019-II190075) funded by the Korea government (MSIT), and NAVER-Intel Co-Lab.
\end{ack}

\clearpage
\bibliographystyle{abbrv}
\bibliography{neurips_2024}
\newpage
\appendix

\section*{\LARGE Appendix}

This document provides the contents that are not included in the main text due to the page limit.

\section{Limitations and Broader Impacts}
\label{appendix:limitation_and_broader_impact}

\paragraph{Limitations}
Despite the promising results, the Meta-Controller framework has several limitations that warrant further investigation.
One significant limitation is its reliance on simulated environments for training and evaluation, which may not fully capture the complexities and variabilities of real-world scenarios.
This gap between simulation and reality could hinder the direct application of the framework to practical robotic tasks.
Additionally, while the framework shows robust performance in the few-shot learning setting, it may still face challenges in environments with highly stochastic dynamics or in tasks that require extensive long-term planning.
Another limitation is the assumption of a unified joint-level representation, which, although effective for the embodiments tested, may not generalize well to robots with significantly different morphologies or actuation mechanisms.
Furthermore, the computational complexity of the transformer-based architecture could pose scalability issues for real-time applications, especially on resource-constrained robotic platforms.
Lastly, the ethical implications of deploying such adaptable robotic systems need to be carefully considered to prevent potential misuse in sensitive areas.

\paragraph{Broader Impacts}
The proposed \modelname{} framework for few-shot imitation learning has several significant broader impacts.
Firstly, its ability to generalize across various robot embodiments and tasks with minimal demonstrations can greatly enhance the adaptability and versatility of robotic systems in dynamic environments.
This flexibility is crucial for deploying robots in real-world scenarios where they must quickly learn and adapt to new tasks without extensive retraining.
Secondly, by leveraging a modular approach that decouples embodiment-specific and task-specific knowledge, the framework promotes efficient knowledge transfer, potentially reducing the computational resources and time required for training new robotic tasks.
However, the deployment of such adaptive systems also raises concerns about their potential misuse.
For instance, advanced robotic systems equipped with this technology could be employed in surveillance or military applications, leading to ethical and privacy issues.
Therefore, it is essential to implement safeguards and ethical guidelines to ensure the responsible use of these advancements.
Furthermore, the reliance on simulated environments for training and evaluation might limit the transferability of the results to real-world conditions, necessitating further research to bridge this gap.
\section{More Details on Experiments}
\label{appendix:more_details}
In this section, we describe the details of the tasks and embodiments in the used dataset and its collection processes.  
\subsection{More details on Datasets and Environment}
\label{appendix:data_details}

\paragraph{Embodiments and Tasks in DeepMind Control (DMC) suite}
The DeepMind Control (DMC) suite benchmark~\cite{tunyasuvunakool2020dmc} is a collection of continuous control tasks implemented using the MuJoCo physics engine~\cite{todorov2012mujoco}. It provides a variety of simulated environments to test and develop reinforcement learning algorithms.
The environments in DMC are designed to be diverse and challenging, promoting the development of agents capable of handling a wide range of tasks.
To reduce the complexity of the problem and facilitate knowledge acquisition from diverse tasks and embodiments, we only consider embodiments that operate on a 2D coordinate space.
We then augment the dataset with a number of tasks used in \cite{wang2018nervenet,hansen2023td}, including two tasks from newly added embodiments, \texttt{walk} and \texttt{run}) task of \texttt{wolf}, where we use the same reward function as the \texttt{walk} and \texttt{run}) task of \texttt{walker}, respectively.
From a total of 38 tasks from 13 embodiments, we select an unseen embodiment with seen tasks (\texttt{reacher-four}, \texttt{wolf}), a seen embodiment with an unseen task (\texttt{walker}), and an unseen embodiment with unseen tasks (\texttt{hopper}) as held-out evaluation tasks for comprehensive analysis.
For the exact setup of the newly added embodiments and tasks, please refer to the github repository of NerveNet~\footnote{\url{https://github.com/WilsonWangTHU/NerveNet}} and TD-MPC2~\footnote{\url{https://github.com/nicklashansen/tdmpc2/tree/main}}.
We list all embodiments and tasks used in our experiments in Table~\ref{tab:all_tasks}.
\begin{table*}[!t]
\caption{List of all embodiments and tasks used in our experiments.}
\label{tab:all_tasks}
\begin{center}
    \renewcommand{\arraystretch}{1.2}
    \renewcommand{\aboverulesep}{0pt}
    \renewcommand{\belowrulesep}{0pt}
    \setlength\tabcolsep{6pt}
    \small
    \begin{tabular}{c | c}
        \toprule
        Embodiment ($\mathcal{E}$) & Task ($\mathcal{T}$)\\
        \midrule
        
        \texttt{acrobot} & \texttt{swingup} \\
        \midrule
        \multirow{2}{*}{\texttt{ball in cup}} & \texttt{catch} \\
        & \texttt{spin} \\
        
        \midrule
        \multirow{2}{*}{\texttt{cartpole}} & \texttt{balance} \\
                      & \texttt{swingup} \\
        \midrule
        \texttt{cartpole two} & \texttt{poles} \\
        \midrule
        \multirow{10}{*}{\texttt{cheetah}} & \texttt{flip} \\
         & \texttt{flip backwards} \\
         & \texttt{jump} \\
         & \texttt{legs up} \\
         & \texttt{lie down} \\
         & \texttt{run back} \\
         & \texttt{run backwards} \\
         & \texttt{run front} \\
         & \texttt{stand back} \\
         & \texttt{stand front} \\
        \midrule
        \multirow{3}{*}{\texttt{hopper}} & \texttt{hop} \\
         & \texttt{hop backwards} \\
         & \texttt{stand} \\
        \midrule
        \multirow{2}{*}{\texttt{pendulum}} & \texttt{spin} \\
         & swingup \\
        \midrule
        \texttt{pointmass} & \texttt{easy} \\
        \midrule
        \texttt{reacher} & \texttt{hard} \\
        \midrule
        \multirow{2}{*}{\texttt{reacher four}} & \texttt{easy} \\
         & \texttt{hard} \\
        \midrule
        \texttt{reacher three} & \texttt{hard} \\
        \midrule
        \multirow{10}{*}{\texttt{walker}} & \texttt{arabesque} \\
         & \texttt{backflip} \\
         & \texttt{flip} \\
         & \texttt{headstand} \\
         & \texttt{legs up} \\
         & \texttt{lie down} \\
         & \texttt{run} \\
         & \texttt{stand} \\
         & \texttt{walk} \\
         & \texttt{walk backwards} \\
        \midrule
        \multirow{2}{*}{\texttt{wolf}} & \texttt{run} \\
        & \texttt{walk} \\
        \bottomrule
    \end{tabular}
    \vspace{-0.5cm}
\end{center}
\end{table*}

\paragraph{Dataset Collection}
For each task, we train the DrQ-v2 agent~\cite{yarats2021image} using an online replay buffer. 
The agent is trained with the hyperparameters specified in \cite{yarats2021image}, generating 1000 to 2000 demonstration trajectories for each task.
For the \texttt{wolf} embodiment that was not originally included in the experiments of the DrQ-v2, we use the same hyperparameters as those used for the corresponding tasks of \texttt{walker}.
After training, we filter out low-quality demonstrations with cumulative rewards smaller than $10$ to ensure the reliability of the training data.

\paragraph{Data Processing}
The states of various embodiments and tasks in DMC consist of two joint-level observations, such as the relative position and velocity of each joint, as well as extrasensory observations that are required to solve each task (\emph{e.g.,} in \texttt{reacher} embodiment, the distance between its tip position and the goal position is given as an extrasensory observation).
The other per-joint attributes such as the movement axis or motion type can be extracted by the kinematic tree of each embodiment provided in XML format.
To ensure compatibility in practical applications, we use only the states defined in the default DMC setting.
There are two different types of joints in the embodiments we consider: hinge joints and slide joints.
Since these joints have distinct motion characteristics (angular motion and linear motion, respectively), we encode the states of each type separately to avoid semantic conflict.
We standardize the relative position using a fixed axis (\emph{e.g.,} the z-axis in the xz plane) instead of using raw, unstandardized input.
For joints with partially existing states (e.g., the free joint of the torso), we zero-pad the non-existent states.
To handle extrasensory states provided by a specific embodiment, we concatenate all extrasensory observations into a single vector and project them into a single joint token $\mathbf{g}_t \in \mathbb{R}^{d}$.
To this end, we introduce embodiment-specific linear projection layers for the embodiments having extrasensory observations.
Lastly, for each embodiment, we normalize the states by scaling them with min-max statistics calculated across the data for each embodiment.

\subsection{More Details on Implementation}
\label{appendix:implementation_details}

\paragraph{Episode Sampling Strategy}
In the episode sampling procedure in Eq.~\eqref{eqn:episodic_meta_learning}, it is crucial to ensure consistency between the query $\mathcal{Q}$ and the few-shot demonstration $\mathcal{D}$ for effective learning from demonstrations.
Since we use a replay buffer as our meta-training dataset which encompasses demonstrations obtained by learning agents at diverse stages, naively applying a uniform sampling strategy on the entire replay buffer results in a high probability of selecting inconsistent queries and demonstrations.
Instead, to ensure task consistency in each episode during episodic meta-learning (Section~\ref{sec:training}), we sample the conditioning demonstrations $\mathcal{D}$ and the query data $\mathcal{Q}$ for Eq.~\eqref{eqn:episodic_meta_learning} from a temporal segment of each replay buffer $\mathcal{B}_{(\mathcal{E}, \mathcal{T})}$.
In other words, the demonstrations in $\mathcal{D}$ and $\mathcal{Q}$ are obtained from expert agents at adjacent training epochs.
In all experiments, we use a temporal segment size of 10 for episodic meta-learning.

\paragraph{Training Details}
All models are trained for 200,000 iterations using the Adam optimizer~\cite{kingma2014adam} and a \textit{poly} learning rate scheduler~\cite{liu2015parsenet} with a base learning rate of $2 \times 10^{-4}$.
After training, we fine-tune all models for 10,000 iterations with a fixed learning rate of $2 \times 10^{-4}$, except for HDT~\cite{xu2022hyper}, which requires a higher learning rate of $10^{-2}$.
For meta-training, we train the model with $8 \times \text{RTX A6000}$ GPUs for approximately 25 hours, and we fine-tune the model on each task with $1 \times \text{RTX A6000}$ GPU for approximately 2 hours. 

As the baselines used in our experiments have not been demonstrated in our few-shot behavior cloning setting with unseen embodiments and tasks, we modify their base architecture or learning framework for fair comparison.
When fine-tuning HDT, we first adapt the embodiment-specific head parameters of HDT with the episodic meta-learning objective, as it does not generate appropriate adapter parameters with untrained head parameters of novel embodiment.
We then obtain adapter parameters by forwarding every temporal chunk of the few-shot demonstration with the hyper-network and initialize adapter parameters by the mean of the outputs.
Then, following the HDT procedure, we fine-tune only the adapter parameters using the BC objective.
For fine-tuning the PDT+PEFT baseline, we apply PEFT techniques to the PDT model meta-trained without task-specific parameters.
For the few-shot adaptation of L2M, the learnable modulation pool and the corresponding low-rank projection matrices are fine-tuned.

\paragraph{Architectural Details}
\begin{table*}[!t]
\caption{Hyper-parameters of \modelname{} used in our experiments.}
\label{tab:hyperparameters}
\begin{center}
    \renewcommand{\arraystretch}{1.2}
    \renewcommand{\aboverulesep}{0pt}
    \renewcommand{\belowrulesep}{0pt}
    \setlength\tabcolsep{6pt}
    \small
    \begin{tabular}{c | c}
        \toprule
        Hyper-parameters & Value \\
        \midrule
        Number of demonstrations used in each episode  & 4 \\
        Global Batch Size & 64 \\
        Hidden dimension & 512 \\
        Attention heads & 4 \\
        Low-rank for structure encoder $f_s$ & 16 \\ 
        Low-rank for motion encoder $f_m$ & 16 \\
        Layerscale initialization & 1 \\
        Training iteration & 200,000 \\        
        Learning rate warmup iterations & 1000 \\ 
        Base learning rate & $2\times 10^{-4}$ \\
        \bottomrule
    \end{tabular}
\end{center}
\end{table*}
For transformer models, we employ a Pre-LN transformer layer~\cite{xiong2020layer}, which consists of a self-attention layer followed by a 2-layer MLP with $4\times$ hidden dimension size.
We use GELU~\cite{hendrycks2016gaussian} activation for the self-attention layers.
To enhance the expressiveness of our matching-based policy network, we implement the matching module $\sigma$ in Eq~\eqref{eqn:matching} with multi-head cross-attention~\cite{vaswani2017attention}, following Kim~et.~al.~\cite{kim2022universal} and Kim~et.~al.~\cite{kim2024chameleon}.
We list the hyper-parameters of \modelname{} in Table ~\ref{tab:hyperparameters}.

\paragraph{Implementation Framework}
We implemented our model based on PyTorch Lightning~\cite{falcon2019pytorch} which supports both Intel Gaudi-v2 (HABANA) and NVIDIA AI accelerators (CUDA). 
We provide the code for both systems on separate branches in the GitHub repository.
\section{Additional Results}
\label{appendix:additional_results}

In this section, we provide additional results and analysis on computational efficiency and robustness.

\subsection{Additional Results on 3-Shot Behavior Cloning}
\begin{table*}[!t]
\caption{$3$-shot behavior cloning results on DeepMind Control suite.}
\label{tab:threeshot_table}
\begin{center}
    \renewcommand{\arraystretch}{1.2}
    \renewcommand{\aboverulesep}{0pt}
    \renewcommand{\belowrulesep}{0pt}
    \setlength\tabcolsep{4pt}
    \scriptsize
    \begin{tabular}{Rc@{\hskip 0.2cm}c@{\hskip 0.2cm}c|c@{\hskip 0.2cm}c|c@{\hskip 0.2cm}RRc}
        \toprule
        \multirow{1}{*}{} &
        \multicolumn{7}{R}{Unseen Emb.} &
        Seen Emb. & 
        \multirow{3}{*}{Avg.} \\
        \cmidrule{1-9}
        Emb. ($\mathcal{E}$) &
        \multicolumn{3}{c|}{\texttt{hopper}} & 
        \multicolumn{2}{c|}{\texttt{wolf}} & 
        \multicolumn{2}{R}{\texttt{reacher-four}} & 
        \multicolumn{1}{R}{\texttt{walker}} &
        \\

        \cmidrule{1-9}
        
        Task ($\mathcal{T}$) &
        \texttt{hop} &
        \texttt{hop-bwd.} &
        \texttt{stand} &
        \texttt{walk} &
        \texttt{run} &
        \texttt{easy} & 
        \texttt{hard} & 
        \texttt{walk-bwd.} &
        \\
        \midrule
        MT-DT &
        7.9 $\pm$ 1.5 &
        40.0 $\pm$ 8.2 &
        5.6 $\pm$ 0.9 &
        49.8 $\pm$ 9.9 &
        32.2 $\pm$ 7.1 &
        10.1 $\pm$ 5.0 &
        4.5 $\pm$ 3.7 &
        58.2 $\pm$ 6.7 &
        25.2 \\
        
        PDT+PEFT &
        2.5 $\pm$ 1.2 &
        42.6 $\pm$ 7.1 &
        8.3 $\pm$ 2.2 &
        39.0 $\pm$ 10.8 &
        37.9 $\pm$ 5.7 &
        8.3 $\pm$ 5.6 &
        6.7 $\pm$ 5.1 &
        \textbf{60.0 $\pm$ 10.7} &
        25.7 \\
        
        \midrule
        MetaMorph &
        13.4 $\pm$ 2.8 &
        76.8 $\pm$ 2.5 &
        33.9 $\pm$ 4.3 &
        51.8 $\pm$ 8.4 &
        26.4 $\pm$ 2.5 &
        2.5 $\pm$ 3.7 &
        4.0 $\pm$ 4.2 &
        18.8 $\pm$ 8.2 &
        30.1 \\
        
        MTGv2 &
        15.5 $\pm$ 3.0 &
        \textbf{77.4 $\pm$ 2.7} &
        29.0 $\pm$ 2.7 &
        65.0 $\pm$ 5.9 &
        25.1 $\pm$ 4.0 &
        -4.4 $\pm$ 1.9 &
        2.8 $\pm$ 2.4 &
        19.4 $\pm$ 6.6 &
        27.9 \\
        
        \midrule
        
        Ours &
        \textbf{34.8 $\pm$ 4.0} &
        75.3 $\pm$ 5.9 &
        \textbf{68.3 $\pm$ 6.5} &
        \textbf{78.2 $\pm$ 6.0} &
        \textbf{43.6 $\pm$ 4.7} &
        \textbf{51.3 $\pm$ 8.9} &
        \textbf{73.2 $\pm$ 9.1} &
        53.5 $\pm$ 7.8 &
        \textbf{57.0} \\
        
        \bottomrule
    \end{tabular}
\end{center}
\end{table*}

To investigate performance in lower-shot settings, we evaluate our model with 3-shot behavior cloning on tasks presented in table~\ref{tab:main_table}.
As shown in Table~\ref{tab:threeshot_table}, our model consistently outperforms baseline approaches, demonstrating robustness even with fewer demonstrations. 
This result highlights the model’s adaptability to fewer examples while maintaining reliable generalization across different tasks and embodiments.

\subsection{Additional Results on Embodiment Variations}
\begin{table*}[!t]
\caption{$5$-shot behavior cloning results on variations of embodiments in DeepMind Control suite.}
\label{tab:emb_variants}
\begin{center}
    \renewcommand{\arraystretch}{1.2}
    \renewcommand{\aboverulesep}{0pt}
    \renewcommand{\belowrulesep}{0pt}
    \scriptsize
    \begin{tabular}{Rc|c|c|c|c|Rc}
        \toprule
        \cmidrule{1-7}
        Target Env. ($\mathcal{E}, \mathcal{T}$) &
        \multicolumn{4}{c|}{\texttt{hopper}-\texttt{stand}} & 
        \multicolumn{2}{R}{\texttt{wolf}-\texttt{walk}} & 
        \multirow{3}{*}{Avg.} \\

        \cmidrule{1-7}
        
        \multirow{2}{*}{Emb. Variations} &
        \multicolumn{2}{c|}{foot length} & 
        \multicolumn{2}{c|}{calf-thigh ratio} & 
        \multicolumn{2}{R}{front-back leg ratio} & 
        \\

        \cmidrule{2-7}
        &
        50\% &
        200\% &
        2:1 &
        1:2 &
        2:1 &
        1:2 &
        \\
        
        \midrule
        MT-DT &
        7.5 $\pm$ 5.4 &
        23.0 $\pm$ 8.2 &
        2.1 $\pm$ 0.7 &
        11.0 $\pm$ 2.3 &
        73.1 $\pm$ 5.8 &
        52.1 $\pm$ 11.9 &
        28.1 \\
        
        PDT+PEFT &
        3.2 $\pm$ 1.2 &
        14.4 $\pm$ 7.8 &
        4.4 $\pm$ 3.4 &
        15.0 $\pm$ 6.1 &
        65.3 $\pm$ 6.8 &
        47.8 $\pm$ 10.8 &
        25.0 \\
        
        \midrule
        MetaMorph &
        5.8 $\pm$ 1.7 &
        39.9 $\pm$ 9.5 &
        5.7 $\pm$ 4.7 &
        29.0 $\pm$ 6.2 &
        60.3 $\pm$ 7.8 &
        71.2 $\pm$ 7.9 &
        35.3 \\
        
        MTGv2 &
        8.2 $\pm$ 2.2 &
        13.3 $\pm$ 5.1 &
        4.6 $\pm$ 1.6 &
        24.6 $\pm$ 3.9 &
        73.8 $\pm$ 5.5 &
        4.0 $\pm$ 2.1 &
        21.4 \\
        
        \midrule
        
        Ours &
        \textbf{15.7 $\pm$ 6.8} &
        \textbf{43.0 $\pm$ 10.0} &
        \textbf{12.5 $\pm$ 6.9} &
        \textbf{75.3 $\pm$ 6.6} &
        \textbf{90.3 $\pm$ 3.3} &
        \textbf{73.2 $\pm$ 9.7} &
        \textbf{51.7} \\
    
        \bottomrule
    \end{tabular}
\end{center}
\end{table*}

We further assess model performance on six additional embodiments by adjusting physical parameters, such as joint lengths (e.g., foot length of hopper) and ratios among joints (e.g., calf-thigh ratio of hopper, front leg-back leg ratio of wolf).
These modifications can make the tasks harder, as the original embodiments are optimized for specific tasks.
In Table~\ref{tab:emb_variants}, we compare Meta-Controller with the high-performing baselines from Table~\ref{tab:main_table}.
Our model consistently outperforms all baselines in these challenging variants, demonstrating the robustness and adaptability of our approach to variations in embodiment configurations.

\subsection{Additional Qualitative Results}
\label{appendix:qualitative_results}
\begin{figure}[t!]
    \centering
    \includegraphics[width=1.0\linewidth]{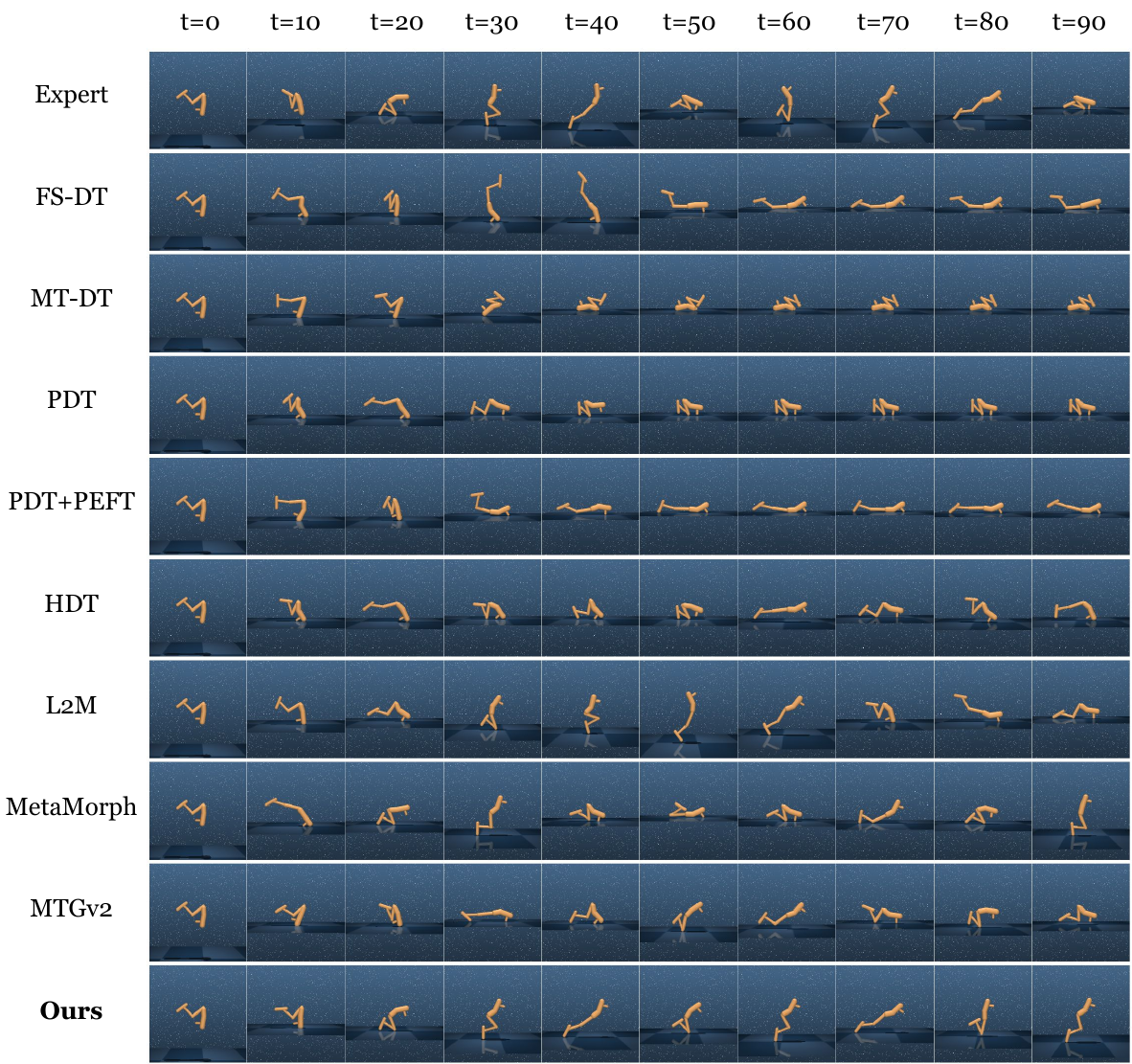}
    \caption{Additional Qualitative Results on \texttt{hopper-hop} task.}
    \vspace{-0.2cm}
    \label{fig:more_qualitative_hopper_hop}
\end{figure}

\begin{figure}[t!]
    \centering
    \includegraphics[width=1.0\linewidth]{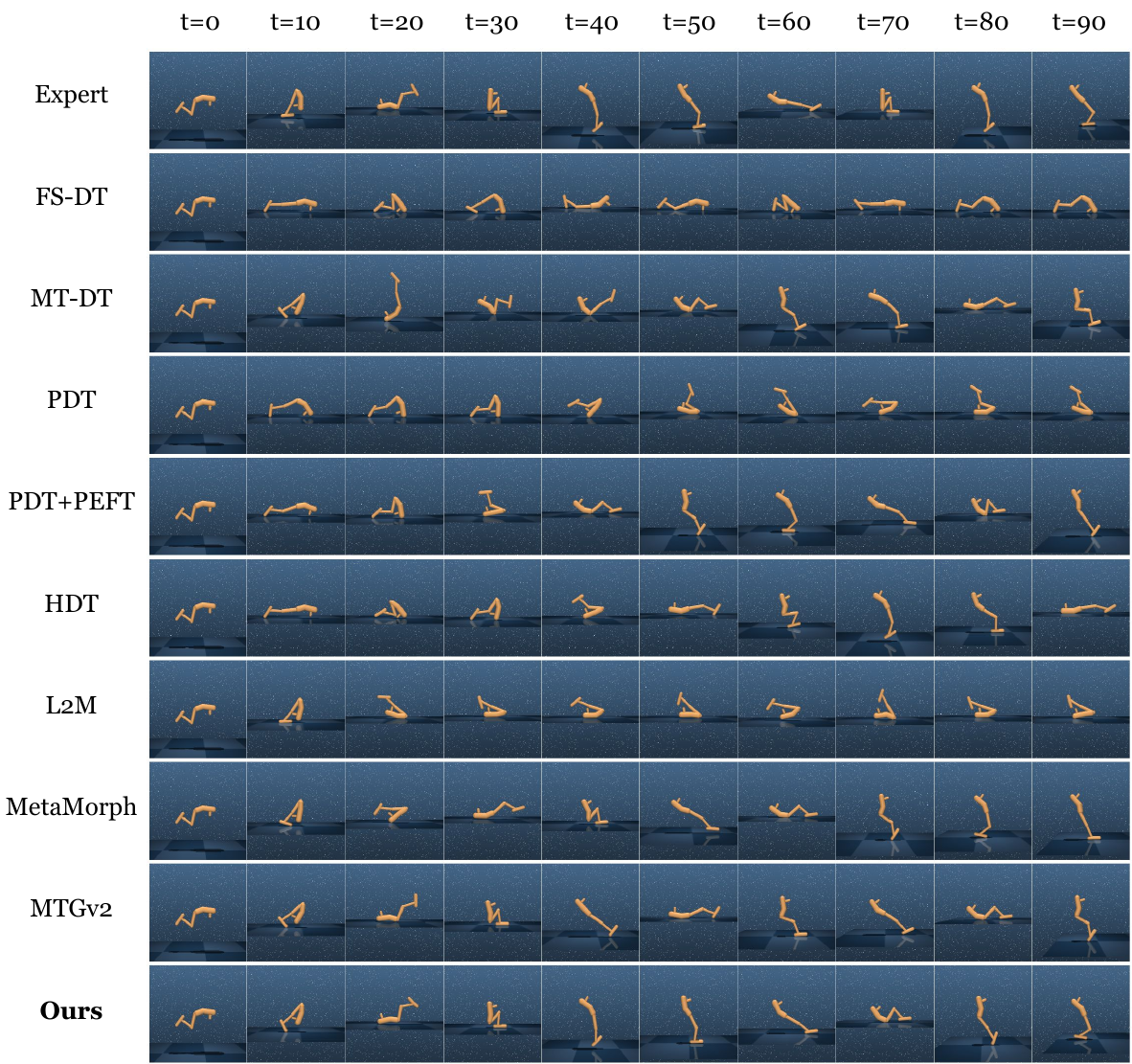}
    \caption{Additional Qualitative Results on \texttt{hopper-hop backwards} task.}
    \vspace{-0.2cm}
    \label{fig:more_qualitative_hopper_hop_bwd}
\end{figure}

\begin{figure}[t!]
    \centering
    \includegraphics[width=1.0\linewidth]{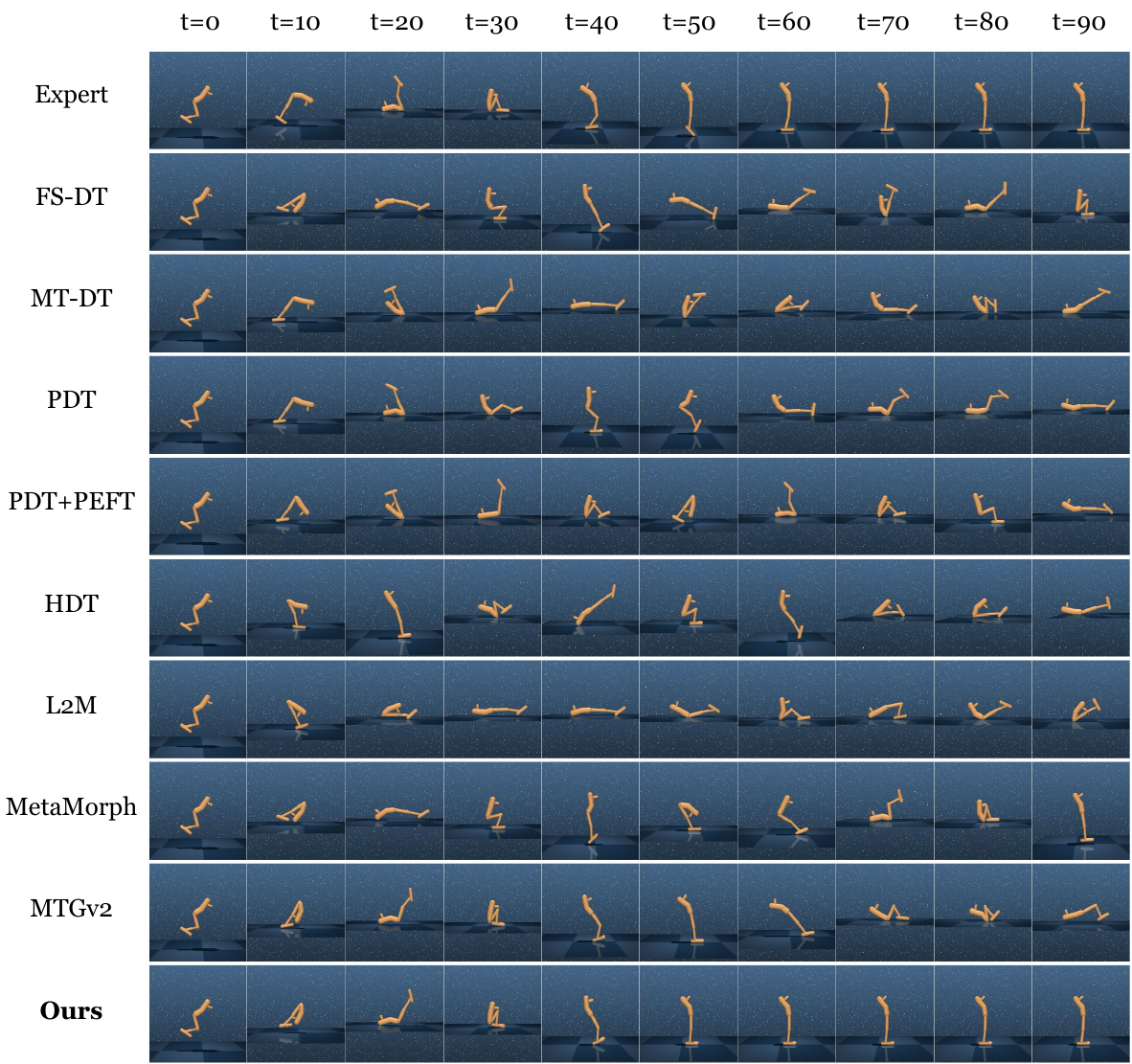}
    \caption{Additional Qualitative Results on \texttt{hopper-stand} task.}
    \vspace{-0.2cm}
    \label{fig:more_qualitative_hopper_stand}
\end{figure}

\begin{figure}[t!]
    \centering
    \includegraphics[width=1.0\linewidth]{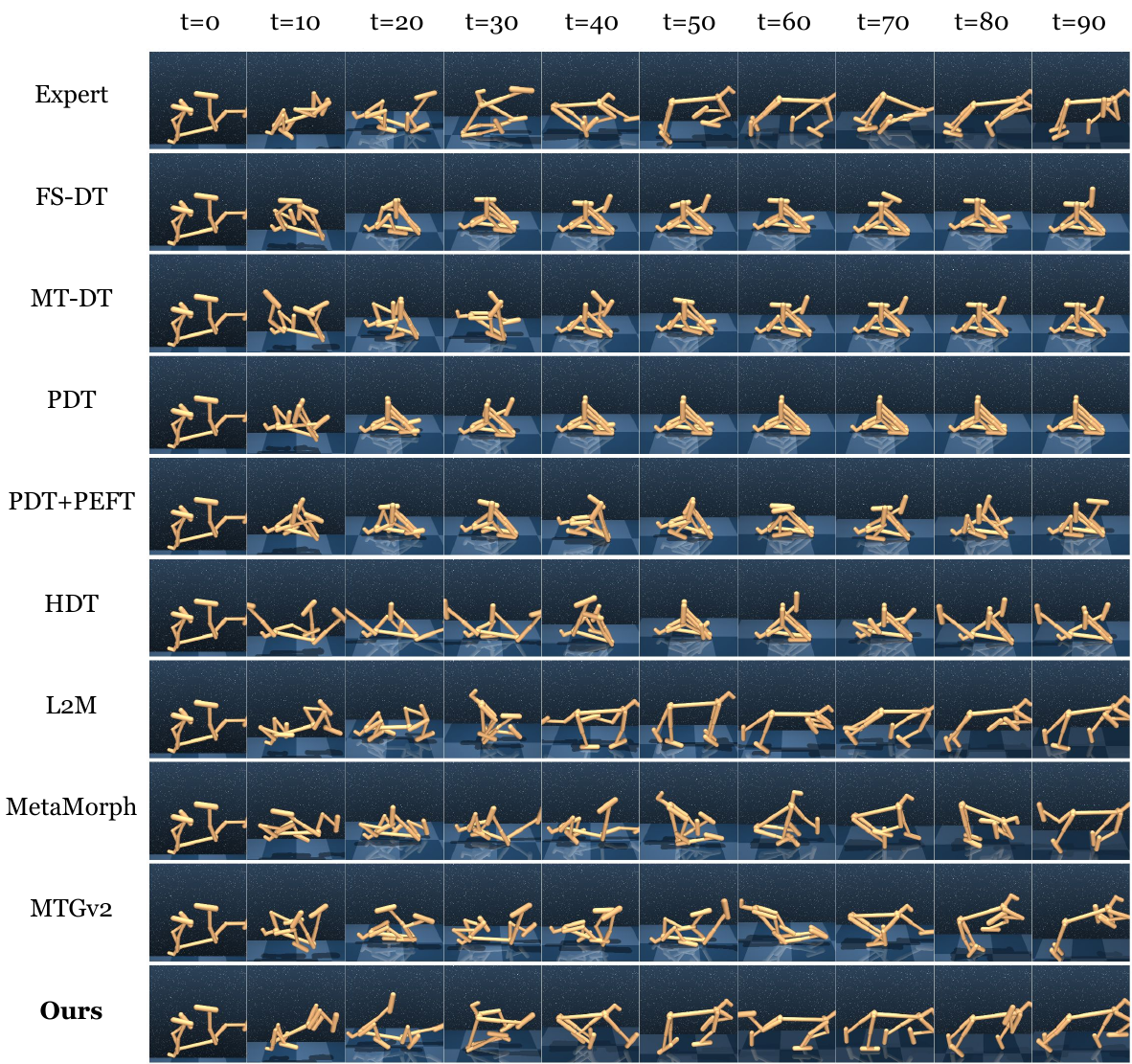}
    \caption{Additional Qualitative Results on \texttt{wolf-walk} task.}
    \vspace{-0.2cm}
    \label{fig:more_qualitative_wolf_walk}
\end{figure}

\begin{figure}[t!]
    \centering
    \includegraphics[width=1.0\linewidth]{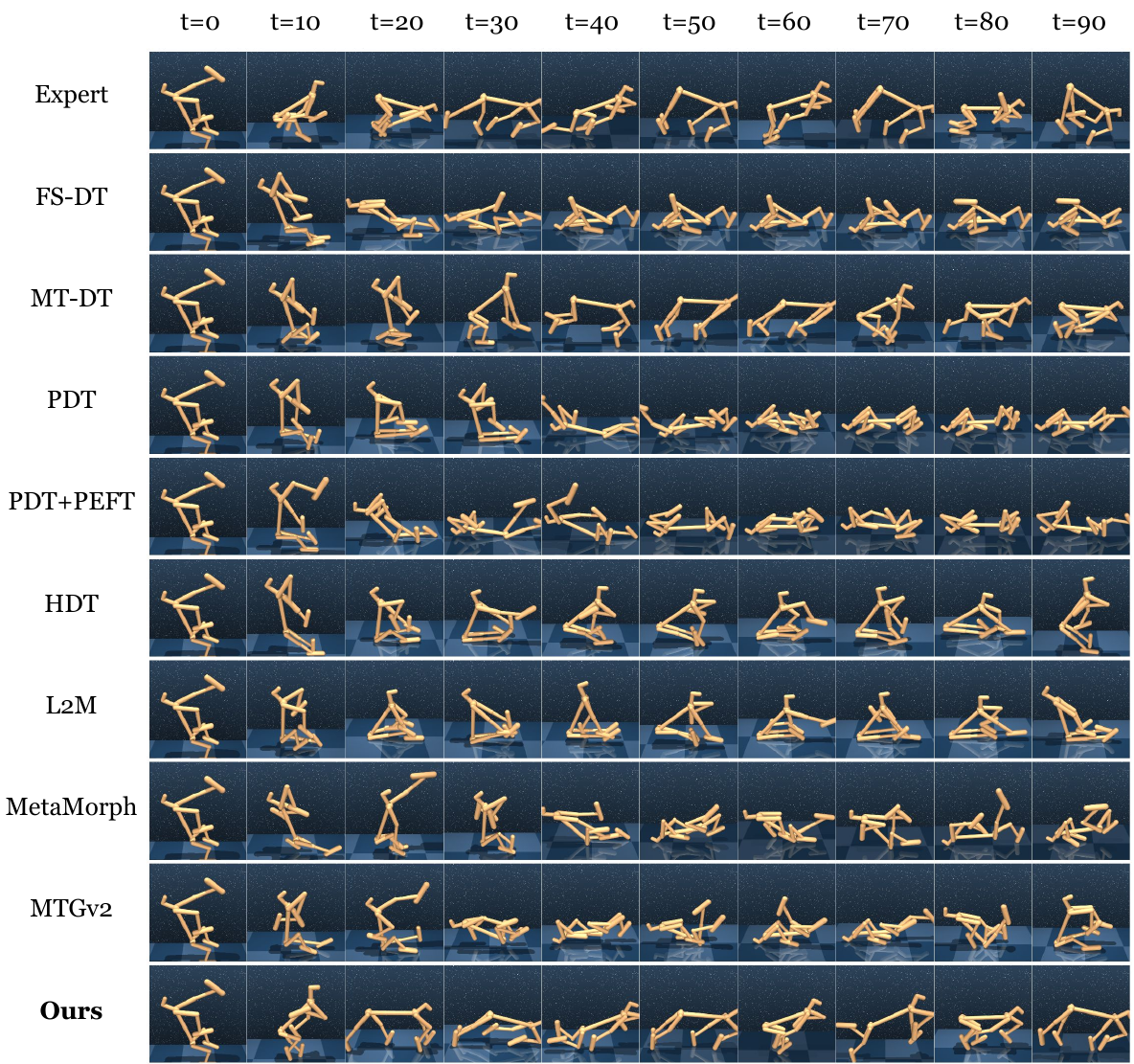}
    \caption{Additional Qualitative Results on \texttt{wolf-run} task.}
    \vspace{-0.2cm}
    \label{fig:more_qualitative_hopper_wolf_run}
\end{figure}

\begin{figure}[t!]
    \centering
    \includegraphics[width=1.0\linewidth]{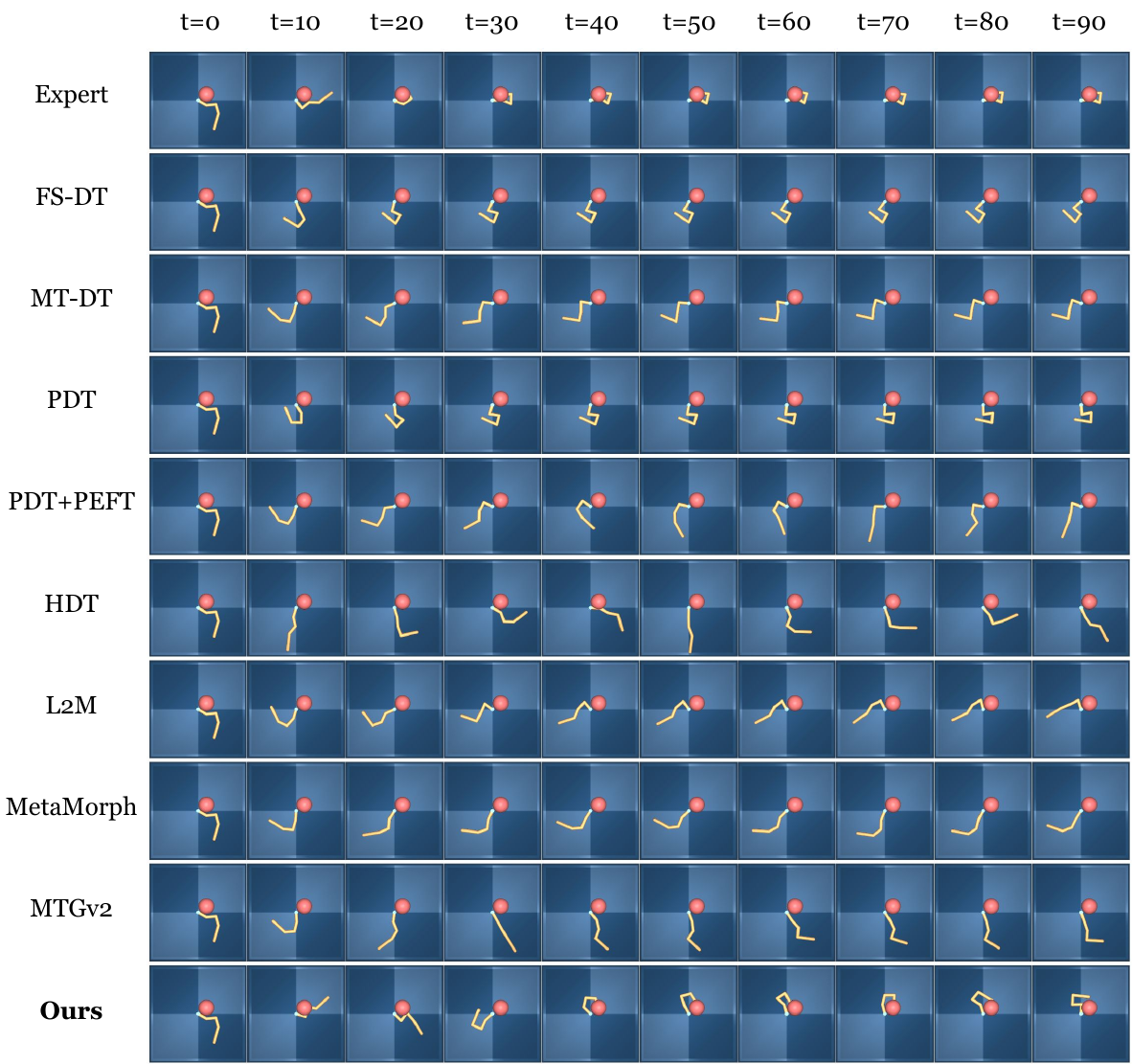}
    \caption{Additional Qualitative Results on \texttt{reacher four-easy} task.}
    \vspace{-0.2cm}
    \label{fig:more_qualitative_reacher_four_easy}
\end{figure}

\begin{figure}[t!]
    \centering
    \includegraphics[width=1.0\linewidth]{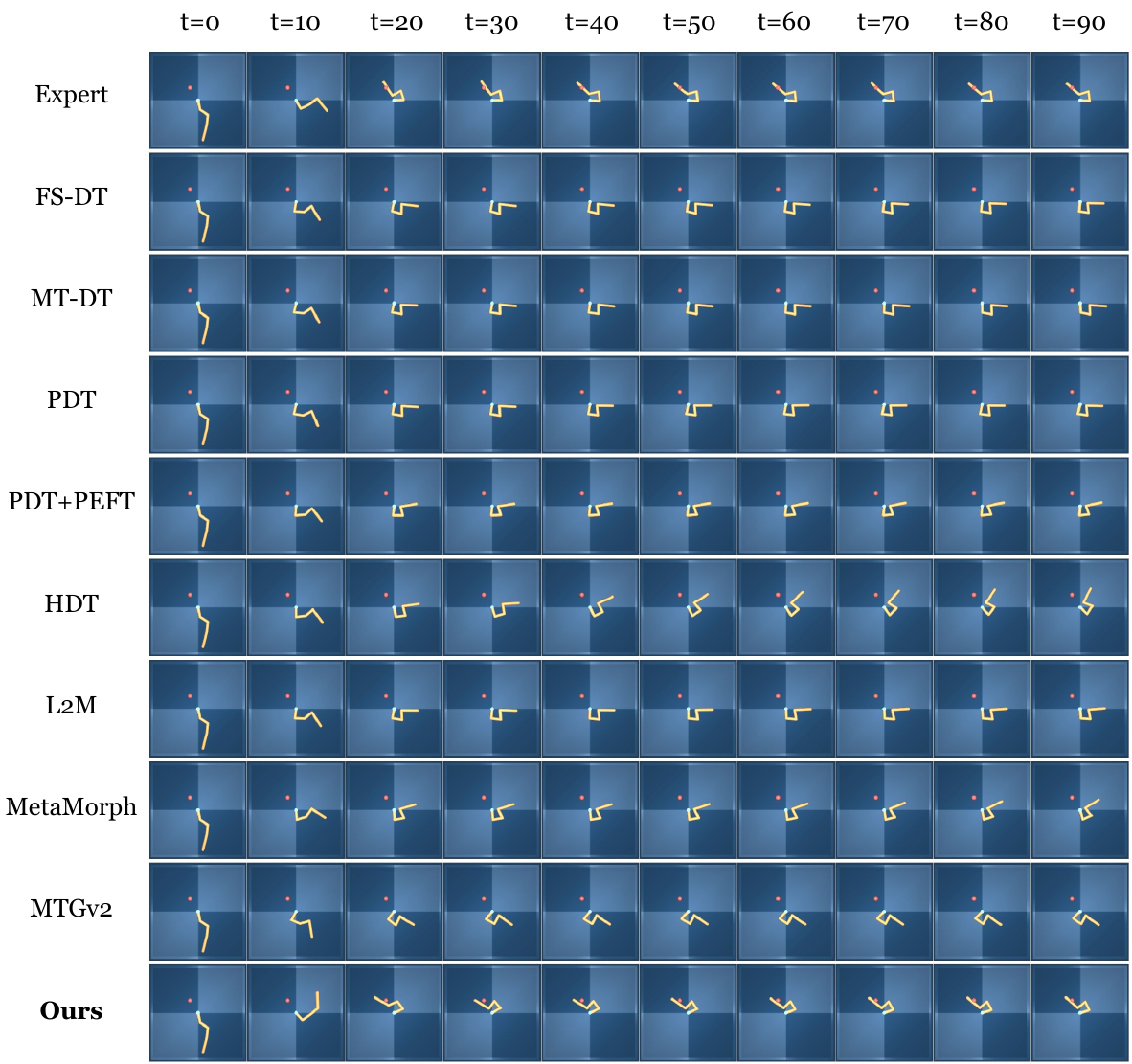}
    \caption{Additional Qualitative Results on \texttt{reacher four-hard} task.}
    \vspace{-0.2cm}
    \label{fig:more_qualitative_reacher_four_hard}
\end{figure}

\begin{figure}[t!]
    \centering
    \includegraphics[width=1.0\linewidth]{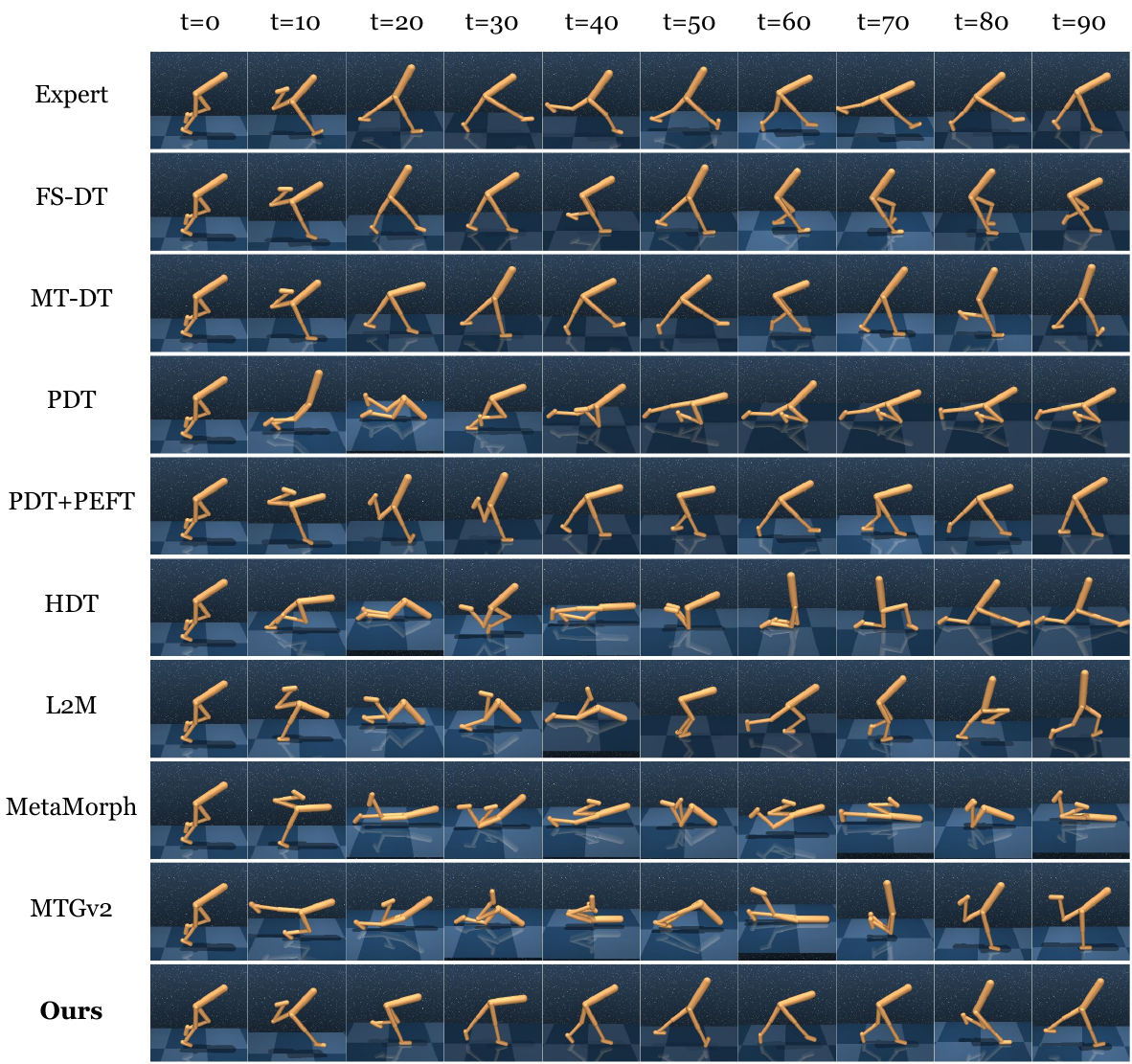}
    \caption{Additional Qualitative Results on \texttt{walker-walk backwards} task.}
    \vspace{-0.2cm}
    \label{fig:more_qualitative_walker_walk_bwd}
\end{figure}

We provide additional qualitative results about the behavior of the few-shot agents in the tasks we used for evaluation.
In Figure~\ref{fig:more_qualitative_hopper_hop}-\ref{fig:more_qualitative_walker_walk_bwd}, we plot the rendered images of a single evaluation rollout by the \texttt{dm\_control} library~\cite{tunyasuvunakool2020dmc} from the initial state ($t=0$) to the states unrolled by each agent, by skipping every ten timesteps in the visualization.

\subsection{Visualization of Learning Curves}
\begin{figure}
    \centering
    \includegraphics[width=\linewidth]{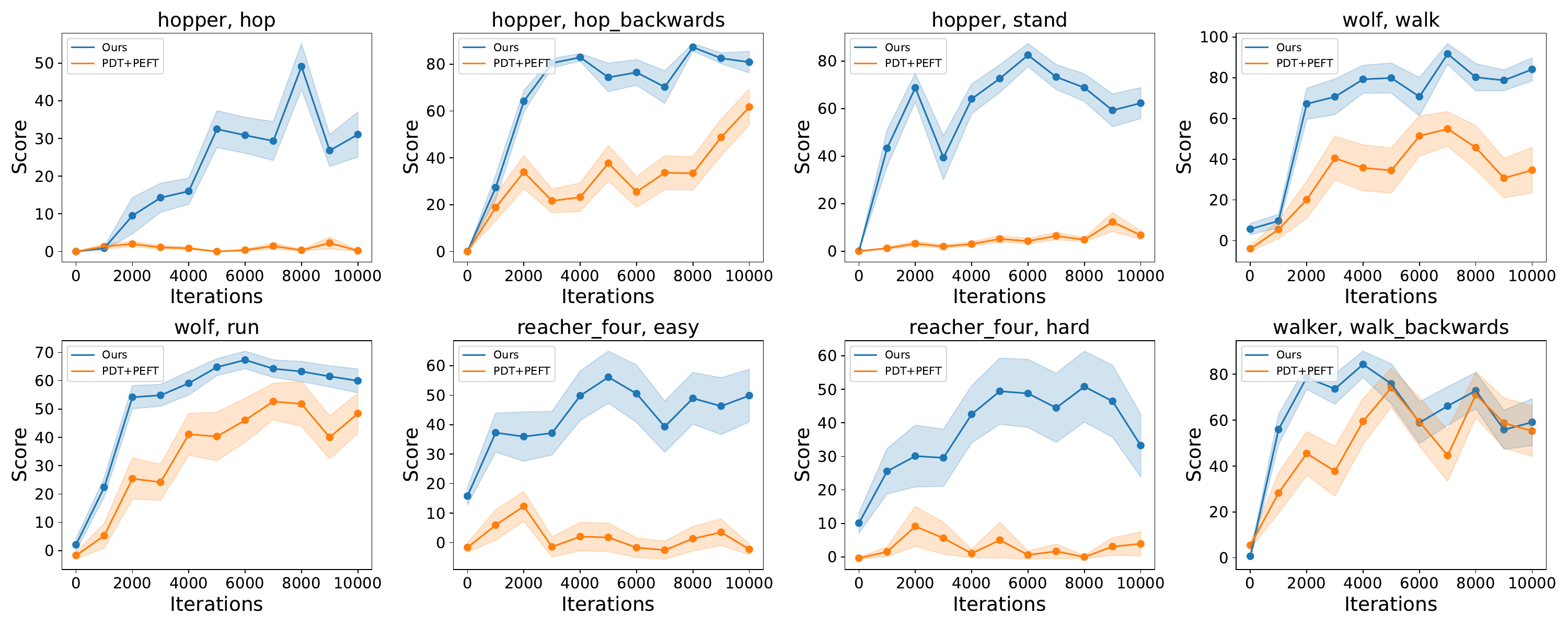}
    \captionof{figure}{Learning curves of ours and PDT+PEFT in $5$-shot settings.}
    \label{fig:learning_curve}
\end{figure}

We provide learning curves of Meta-Controller and PDT+PEFT in Figure~\ref{fig:learning_curve}.
Due to the modular nature of our structure encoder, our model not only achieves better performance but also converges much more quickly than PDT+PEFT in every task.

\subsection{Visualization of Learned Embeddings}
\begin{figure}
\centering
    \begin{subfigure}{.47\textwidth}
      \centering
        \includegraphics[width=\linewidth]{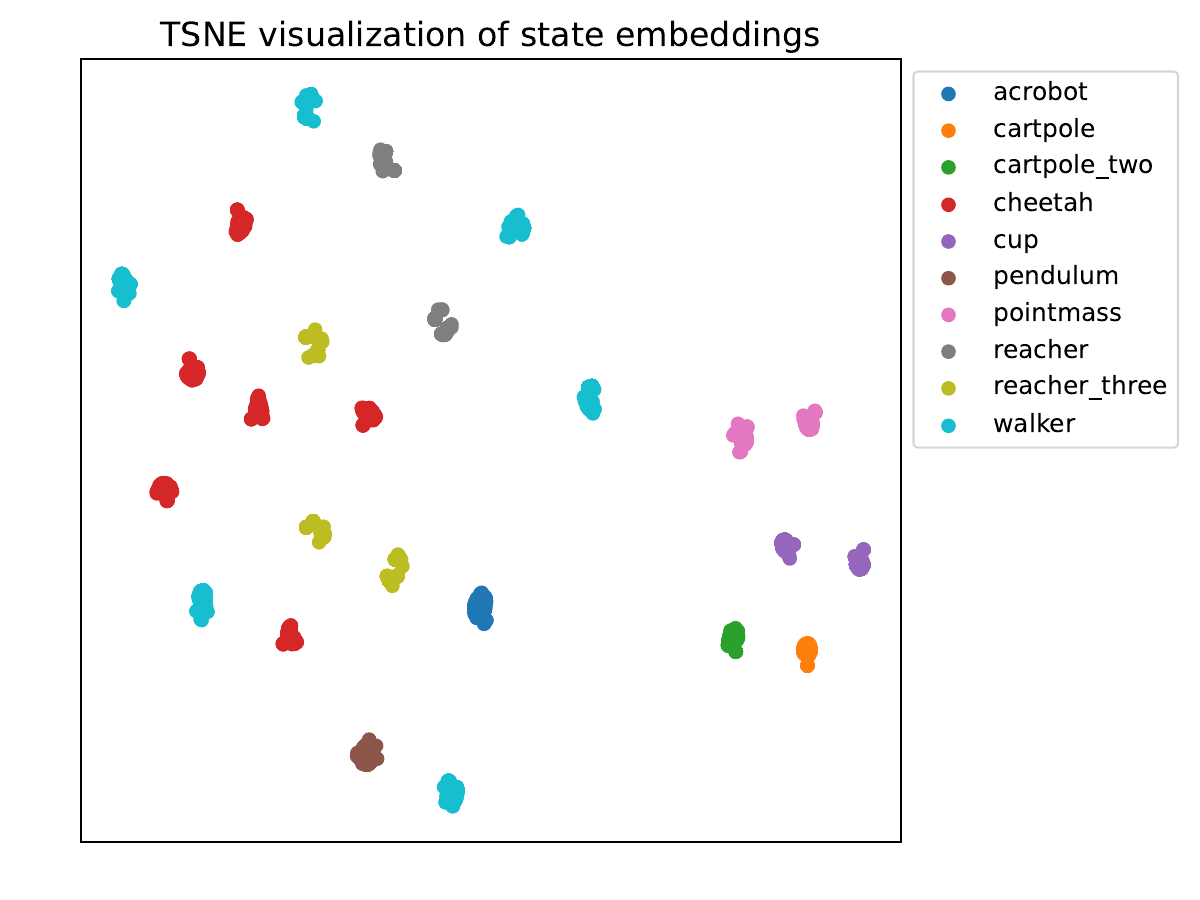}
    \end{subfigure}%
    \begin{subfigure}{.47\textwidth}
      \centering
      \includegraphics[width=\linewidth]{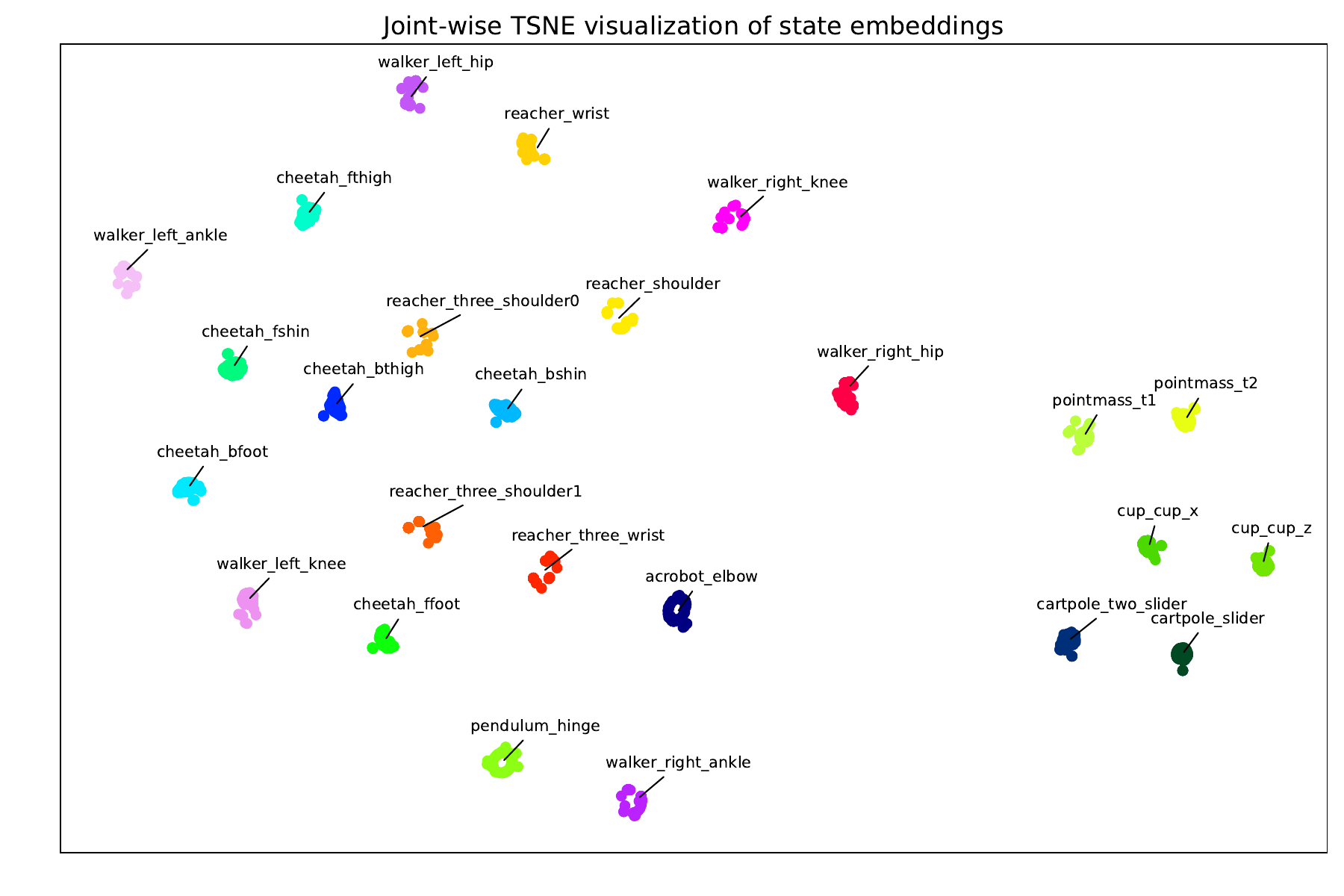}
    \end{subfigure}
    \vspace{-0.2cm}
    \caption{t-SNE visualization of embeddings of state encoder $f$. In the left figure, embeddings of the same embodiment have the same color. In the right figure, embeddings of the same joint have the same color.}
\label{fig:tsne_vis}
\end{figure}

To gain insights into how our model represents different embodiments, we provide a t-SNE visualization of embedding space of features obtained by structure encoder $f_s$ in Figure~\ref{fig:tsne_vis}.
First, we observe that the embeddings are clustered by the joints of each embodiment, and the clusters corresponding the same embodiment are located nearby.
Also, we note that the embeddings of slide joints (e.g., \texttt{cartpole}, \texttt{cartpole-two}, \texttt{cup}, \texttt{pointmass}) and the embeddings of hinge joints (e.g., \texttt{reacher}, \texttt{reacher-three}, \texttt{walker}, \texttt{acrobot}, \texttt{cheetah}, \texttt{pendulum}) are separated in the right and the left regions. 
Thus, the state encoder captures both embodiment-specific and joint-specific knowledge, providing rich features to the policy network.

\subsection{Computational Complexity Analysis}
\begin{table*}[!t]
    \begin{center}
        \caption{Inference time of ours and VC-1~\cite{majumdar2023we}, measured in a single NVIDIA RTX 3090 GPU.}
        \label{tab:inference_time}
        \begin{tabular}{c|c|c}
            \toprule
             Model & Ours & VC-1 (ViT-L backbone) \\
             \midrule
             Inference Time (ms) & 17.8 & 26.8  \\
             \bottomrule
        \end{tabular}
    
        \captionof{table}{Relative inference time of each module of ours, measured in a single NVIDIA RTX 3090 GPU.}
        \label{tab:modulewise_time}
        \begin{tabular}{c|c|c|c|c}
            \toprule
             Module & State encoder ($f$) & Action encoder ($g$) & Action decoder ($h$) & Matching ($\sigma$) \\
             \midrule
             Inference Time (\%) & 31.54 & 31.83 & 34.08 & 2.55 \\
             \bottomrule
        \end{tabular}
    \end{center}
\end{table*}
Table~\ref{tab:inference_time} presents the inference time comparisons between our model and VC-1~\cite{majumdar2023we}, which is a ViT-based foundation model designed for behavior cloning of continuous control tasks.
Despite VC-1’s high performance in visual tasks, its transformer-based architecture incurs notable computational costs.
Our model achieves faster inference times, highlighting its efficiency relative to VC-1.
We also note that models like RT-2~\cite{brohan2023rt}, which are designed for real-time robotic manipulation, typically involve architectures such as a 40-layer ViT and a large language model (LLM) with 3 billion parameters, requiring significantly higher computational resources than our approach.

Additionally, Table~\ref{tab:modulewise_time} provides a breakdown of relative inference times for each module in our model, demonstrating that the majority of computation time is allocated to the transformer-based encoders and decoders.
In this context, advancements in optimizing transformers, such as sparse attention mechanisms~\cite{choromanski2020rethinking}, model pruning~\cite{kim2022learned}, and knowledge distillation~\cite{sanh2019distilbert}, can enhance inference speeds. 
Since these techniques are orthogonal to ours, they can be naturally incorporated into our method for resource-constrained robotic platforms that require high speed inference.

\subsection{Failure Case Analysis}
\begin{figure}
    \centering
    \includegraphics[width=\linewidth]{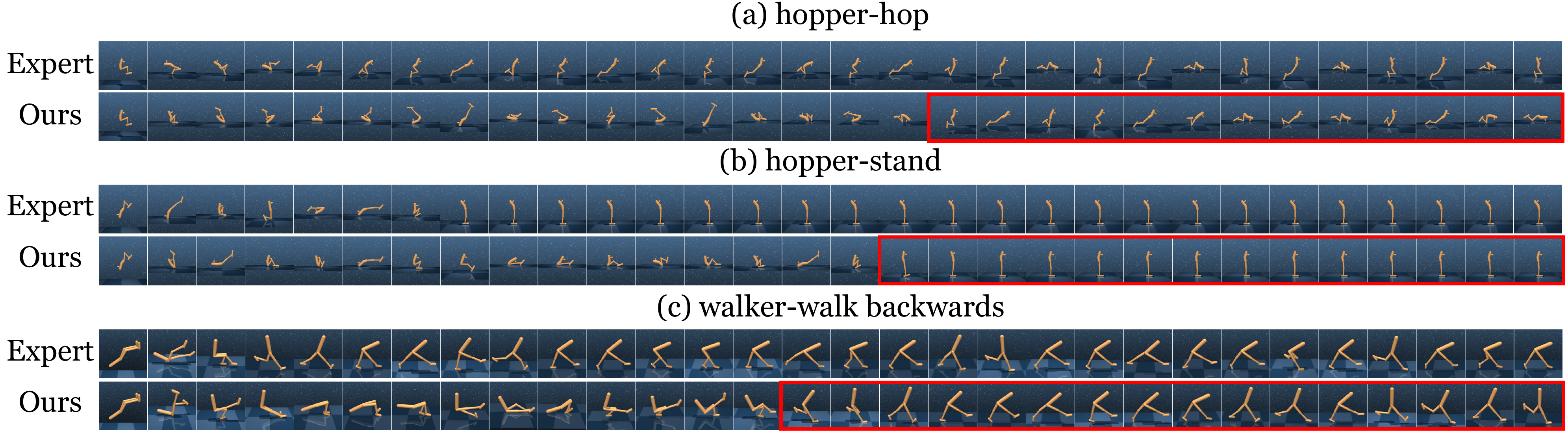}

    \captionof{figure}{Failure cases of our model. As indicated by red boxes, the agent begins to solve tasks effectively after it reaches a specific posture.}
    \label{fig:failure_case}

    \includegraphics[width=0.5\linewidth]{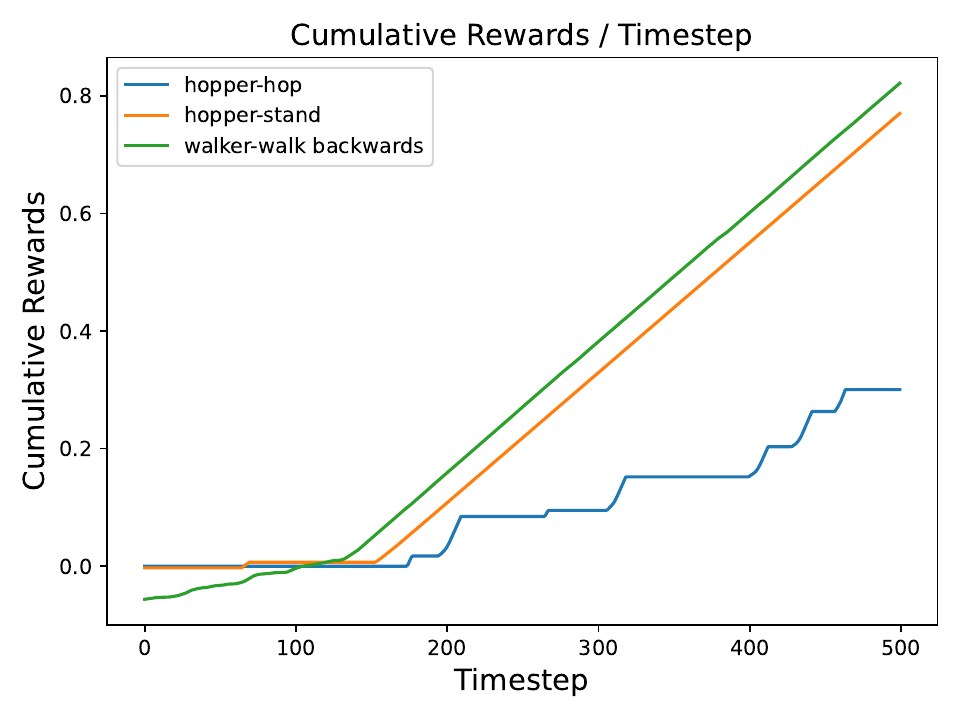}
    \captionof{figure}{Cumulative rewards of the failure cases in Figure~\ref{fig:failure_case}.}
    \label{fig:failure_case_rewards}
\end{figure}

We present visualizations of scenarios where the Meta-Controller performs less effectively and cumulative rewards over time for each scenario in Figures~\ref{fig:failure_case} and \ref{fig:failure_case_rewards}, respectively.
In these failure cases, we observed that agents struggle to obtain rewards until they reach a specific posture. 
Once they achieve this posture (highlighted by the red box in Figure~\ref{fig:failure_case}), they begin to solve the task effectively. 
This pattern is also reflected in Figure~\ref{fig:failure_case_rewards}, where rewards remain near zero until a certain timestep, after which they rise consistently.
This result implies that encouraging the agent to reach states similar to those in the demonstrations (e.g., via exploration strategies) would improve performance in challenging few-shot scenarios.

\subsection{Robustness Analysis under Noise}
\begin{figure}
    \centering
    \includegraphics[width=\linewidth]{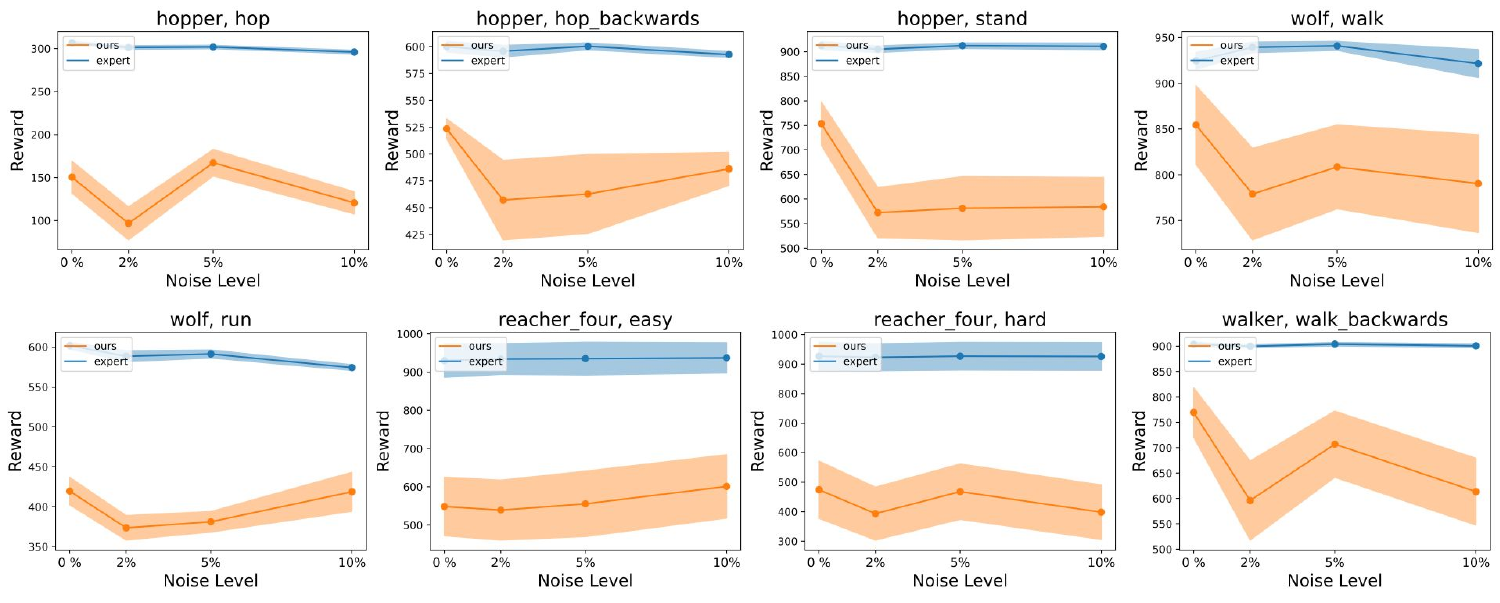}
    \caption{Robustness analysis of our model in the presence of various noise levels in transition dynamics. Random noise $\epsilon \sim \mathcal{U}[-n, n]$ is added to each action, where $n$ varies over 2\%, 5\%, and 10\% of the action range. The shaded region represents the standard error.}
    \label{fig:robustness}
\end{figure}

To assess robustness in noisy environments, we introduce varying levels of noise to the transition dynamics and measure the resulting performance. 
Random noise sampled from $\mathcal{U}[-n, n]$ was added to the agent's action at each timestep, with three noise levels $n \in [2\%, 5\%, 10\%]$ of the action range. 
Figure~\ref{fig:robustness} plots the rewards of our model at each noise level compared to experts.

The results indicate that our method maintains its performance across many tasks as noise levels increase, showing its robustness under stochastic environments.
Interestingly, for tasks such as reacher-four, the performance increases with higher noise levels, likely due to the exploration effect induced by stochastic transitions.
This robustness under stochastic dynamics indicates potential for the model’s application in real-world scenarios where environmental variability is common.

\section{Additional Ablation Studies}
\label{appendix:additional_ablation_studies}

In this section, we provide additional results on ablation studies.

\subsection{Additional Ablation Studies on Architectural Components}
\label{appendix:more_ablation_architecture}
\begin{table*}[!t]
\caption{Additional ablation study on the architectural components. Model variants are evaluated in $5$-shot setting.}
\label{tab:appendix_architectural_compoenents}
\begin{center}
    \renewcommand{\arraystretch}{1.2}
    \renewcommand{\aboverulesep}{0pt}
    \renewcommand{\belowrulesep}{0pt}
    \setlength\tabcolsep{6pt}
    \scriptsize
    \begin{tabular}{@{\hskip 0.1cm}c@{\hskip 0.1cm}c@{\hskip 0.1cm}c@{\hskip 0.1cm}c@{\hskip 0.1cm}Rc@{\hskip 0.1cm}c@{\hskip 0.1cm}c|c@{\hskip 0.1cm}c|c@{\hskip 0.1cm}RRc}
        \toprule
        \multirow{3}{*}{$f_s$} &
        \multirow{3}{*}{$f_m$} &
        \multirow{3}{*}{$\sigma$} &
        \multirow{3}{*}{$g$} &
        \multirow{3}{*}{$h$} &
        \multicolumn{7}{R}{Unseen Emb.} & Seen Emb. & 
        \multirow{3}{*}{Avg.} \\
        \cmidrule{6-13}
        
        & & & & & \multicolumn{3}{c|}{\texttt{hopper}} & 
        \multicolumn{2}{c|}{\texttt{wolf}} & 
        \multicolumn{2}{R}{\texttt{reacher-four}} & 
        \multicolumn{1}{R}{\texttt{walker}} & \\
        \cmidrule{6-13}
        
        & & & & &
        \texttt{hop} &
        \texttt{hop-bwd.} &
        \texttt{stand} &
        \texttt{walk} &
        \texttt{run} &
        \texttt{easy} & 
        \texttt{hard} & 
        \texttt{walk-bwd.} &
        \\

        \midrule
        
         \xmark & \xmark &\xmark & \xmark & \xmark &
        6.4 $\pm$ 2.4 &
        58.2 $\pm$ 8.4 &
        7.7 $\pm$ 1.1 &
        32.4 $\pm$ 8.7 &
        27.4 $\pm$ 6.4 &
        14.6 $\pm$ 7.5 &
        4.9 $\pm$ 4.1 &
        0.2 $\pm$ 1.7 &
        19.0 \\

        \xmark & \cmark & \cmark & \cmark & \cmark &
        5.7$\pm$2.7 &
        44.7$\pm$10.3 &
        15.5$\pm$3.9 &
        64.5$\pm$9.9 &
        53.3$\pm$8.5 &
        9.3$\pm$7.0 &
        0.0$\pm$0.8 &
        48.4$\pm$10.4 &
        30.2 \\
        
         \cmark & \cmark &\cmark & \xmark & \xmark &
        25.7 $\pm$ 5.5 &
        \textbf{89.5 $\pm$ 1.4} &
        42.5 $\pm$ 9.3 &
        83.9 $\pm$ 6.1 &
        60.7 $\pm$ 3.4 &
        53.7 $\pm$ 8.9 &
        \textbf{53.2 $\pm$ 10.7} &
        71.6 $\pm$ 6.0 &
        60.1 \\

        \midrule
        
        \cmark & \cmark & \cmark & \cmark & \cmark &
        \textbf{49.1$\pm$6.1} &
        87.2$\pm$1.6 &
        \textbf{82.5$\pm$4.9} &
        \textbf{91.7$\pm$5.1} &
        \textbf{67.3$\pm$3.1} &
        \textbf{56.1$\pm$8.8} &
        50.8$\pm$10.6 &
        \textbf{84.3$\pm$5.7} &
        \textbf{71.1} \\

        \bottomrule
    \end{tabular}
\end{center}
\end{table*}

Table~\ref{tab:appendix_architectural_compoenents} presents the additional ablation studies on key architectural components:structure encoder $f_s$, motion encoder $f_m$, action encoder $g$, action decoder $h$ and matching module $\sigma$.
Consistent with the discussion in Section~\ref{sec:ablation_study}, we observe that removing the structure encoder $f_s$ (row 2) significantly reduces performance on both seen and unseen embodiments.
This indicates that the structure encoder captures essential morphology-related knowledge that enables cross-embodiment generalization. 

We also ablate the action encoder g and action decoder h (row 3), observing that removing them result in decreased adaptability and performance.
The action encoder plays a key role in transforming raw action into a latent representation, which effectively enhancing expressibility of the policy network and enables adaptation to various unseen tasks with non-convex relationships between states and actions.
Additionally, by encoding actions along the temporal axis, the model can construct a pool of transferrable action features related to local motor skills, which facilitates efficient transfer to unseen tasks that share modular skills but involve different skill combinations.

\subsection{Additional Ablation studies on Adaptation Mechanism}
\label{appendix:more_ablation_adaptation}
\begin{table*}[!t]
\caption{Ablation study on the adaptive parameters in the state encoder \emph{without} the matching module in the policy network.}
\label{tab:tsparam_ablation_womatching}
\begin{center}
    \renewcommand{\arraystretch}{1.2}
    \renewcommand{\aboverulesep}{0pt}
    \renewcommand{\belowrulesep}{0pt}
    \setlength\tabcolsep{6pt}
    \scriptsize
    \begin{tabular}{@{\hskip 0.1cm}c@{\hskip 0.2cm}Rc@{\hskip 0.2cm}c@{\hskip 0.2cm}c|c@{\hskip 0.2cm}c|c@{\hskip 0.2cm}RRc}
        \toprule
        \multirow{3}{*}{$\mathbf{p}_s^{\mathcal{E}}, \theta_s^\mathcal{E}$} &
        \multirow{3}{*}{$\theta_m^{(\mathcal{E}, \mathcal{T})}$} &
        \multicolumn{7}{R}{Unseen Emb.} &
        Seen Emb. &
        \multirow{3}{*}{Avg.} \\
        \cmidrule{3-10}
        
        & & \multicolumn{3}{c|}{\texttt{hopper}} & 
        \multicolumn{2}{c|}{\texttt{wolf}} & 
        \multicolumn{2}{R}{\texttt{reacher-four}} & 
        \multicolumn{1}{R}{\texttt{walker}} &
        \\

        \cmidrule{3-10}
        
        & & 
        \texttt{hop} &
        \texttt{hop-bwd.} &
        \texttt{stand} &
        \texttt{walk} &
        \texttt{run} &
        \texttt{easy} & 
        \texttt{hard} & 
        \texttt{walk-bwd.} &
        \\
        \midrule

        \xmark & \hspace{-0.3cm} \cmark \hspace{-0.3cm} &
        22.6$\pm$3.0 &
        68.1$\pm$3.0 &
        48.1$\pm$5.7 &
        71.7$\pm$4.7 &
        32.6$\pm$3.9 &
        13.7$\pm$6.5 &
        10.3$\pm$5.2 &
        53.1$\pm$9.2 &
        40.0 \\
        
        \cmark & \hspace{-0.3cm} \xmark \hspace{-0.3cm} &
        \textbf{40.0$\pm$5.8} &
        \textbf{91.6$\pm$0.8} &
        53.2$\pm$9.8 &
        78.3$\pm$8.4 &
        \textbf{61.8$\pm$5.1} &
        \textbf{78.1$\pm$7.7} &
        \textbf{36.1$\pm$8.7} &
        2.3$\pm$1.8 &
        55.2 \\
        \midrule
        
        \cmark & \hspace{-0.3cm} \cmark \hspace{-0.3cm} &
        22.8$\pm$5.9 &
        86.3$\pm$1.6 &
        \textbf{83.4$\pm$4.2} &
        \textbf{79.7$\pm$6.4} &
        57.3$\pm$6.7 &
        54.9$\pm$9.5 &
        24.8$\pm$8.2 &
        \textbf{73.4$\pm$5.0} &
        \textbf{60.3} \\
        \bottomrule
    \end{tabular}

\end{center}
\end{table*}
To further analyze the adaptation mechanism of our model, we conduct an additional ablation study by ablating the adaptive parameters in the state encoder \emph{without} using the matching module in the policy network.
On average, using both adaptive parameters (row 3) achieves the best performance.
However, we note that the model variant without adaptive parameters in the motion encoder (row 2) achieves higher performance than using the parameters (row 3) in many tasks.
We conjecture that this is due to over-fitting on the few-shot demonstrations, as the adaptive parameters in the motion encoder are both specific to the embodiment and task and more prone to over-fitting.
However, as we discussed with Table~\ref{tab:tsparam_ablation} in Section~\ref{sec:ablation_study}, such trends are not found in general (except for the \texttt{easy} task of \texttt{reacher-four}).
This indicates that in few-shot learning settings, the PEFT techniques must be employed together with a robust architecture such as matching to exploit its effectiveness maximally.

\subsection{Additional Ablation Studies on Meta-Training Task Composition}
\label{appendix:more_ablation_task_composition}
\begin{table*}[!t]
\caption{Ablation study on embodiments included in the meta-training data. We report the $5$-shot score by removing four embodiments from the meta-training data, where we choose three different subsets of the removing embodiments. $\mathcal{E}_1$, $\mathcal{E}_2$, $\mathcal{E}_3$ denote $\{\texttt{cartpole-two}, \texttt{cup}, \texttt{pendulum}, \texttt{pointmass}\}$, \{\texttt{cartpole-two}, \texttt{pendulum}, \texttt{pointmass}, \texttt{reacher-three}\}, and \{\texttt{cartpole-two}, \texttt{cheetah}, \texttt{cup}, \texttt{reacher-three}\}, respectively.}
\label{tab:mt_composition}
\begin{center}
    \renewcommand{\arraystretch}{1.2}
    \renewcommand{\aboverulesep}{0pt}
    \renewcommand{\belowrulesep}{0pt}
    \setlength\tabcolsep{1.5pt}
    \scriptsize
    \begin{tabular}{Rc@{\hskip 0.2cm}c@{\hskip 0.2cm}c|c@{\hskip 0.2cm}c|c@{\hskip 0.2cm}RRc}
        \toprule
        \multirow{1}{*}{} &
        \multicolumn{7}{R}{Unseen Emb.} &
        Seen Emb. & 
        \multirow{3}{*}{Avg.} \\
        \cmidrule{1-9}
        Emb. ($\mathcal{E}$) &
        \multicolumn{3}{c|}{\texttt{hopper}} & 
        \multicolumn{2}{c|}{\texttt{wolf}} & 
        \multicolumn{2}{R}{\texttt{reacher-four}} & 
        \multicolumn{1}{R}{\texttt{walker}} &
        \\

        \cmidrule{1-9}
        
        Task ($\mathcal{T}$) &
        \texttt{hop} &
        \texttt{hop-bwd.} &
        \texttt{stand} &
        \texttt{walk} &
        \texttt{run} &
        \texttt{easy} & 
        \texttt{hard} & 
        \texttt{walk-bwd.}
        \\
        \midrule

        w/o $\mathcal{E}_1$ &
        15.9 $\pm$ 5.5 &
        76.6 $\pm$ 6.8 &
        71.5 $\pm$ 7.1 &
        83.2 $\pm$ 4.1 &
        62.1 $\pm$ 5.0 &
        \textbf{82.1 $\pm$ 7.6} &
        41.8 $\pm$ 9.8 &
        70.7 $\pm$ 6.0 &
        63.0 \\

        w/o  $\mathcal{E}_2$&
        16.3 $\pm$ 6.0 &
        85.2 $\pm$ 2.8 &
        81.6 $\pm$ 7.3 &
        83.9 $\pm$ 5.2 &
        \textbf{70.1 $\pm$ 3.4} &
        5.6 $\pm$ 6.9 &
        4.6 $\pm$ 5.1 &
        78.8 $\pm$ 7.3 &
        53.3 \\

        w/o $\mathcal{E}_3$ &
        26.6 $\pm$ 8.0 &
        \textbf{88.8 $\pm$ 1.1} &
        68.6 $\pm$ 6.1 &
        83.6 $\pm$ 6.4 &
        59.0 $\pm$ 2.9 &
        4.6 $\pm$ 6.5 &
        3.8 $\pm$ 4.1 &
        81.8 $\pm$ 5.1 &
        52.1 \\
        
        
        Use All Embodiments &
        \textbf{49.1$ \pm$ 6.1} &
        87.2 $\pm$ 1.6 &
        \textbf{82.5 $\pm$ 4.9} &
        \textbf{91.7 $\pm$ 5.1} &
        67.3 $\pm$ 3.1 &
        56.1 $\pm$ 8.8 &
        \textbf{50.8 $\pm$ 10.6} &
        \textbf{84.3 $\pm$ 5.7} &
        \textbf{71.1} \\
        
        \bottomrule
    \end{tabular}
\end{center}
\end{table*}

To further understand how the composition of meta-training tasks affects performance, we conduct experiments with varying subsets of embodiments and tasks in Table~\ref{tab:mt_composition}. 
We select 3 combinations of training tasks, where we remove 4 embodiments from the original 10 embodiments. 
Then, we performed 5-shot behavior cloning experiments on the 8 tasks presented in table~\ref{tab:main_table}.

According to Table~\ref{tab:mt_composition}, we observe that using all embodiments outperforms the baselines in most tasks, indicating that a diverse set of embodiments makes the model robust to unseen embodiments.
Our findings also show that it is crucial to include embodiments with morphology and dynamics similar to the downstream tasks in the meta-training dataset.
For example, removing the \texttt{reacher-three} task (as seen in row 2 and row 3) significantly drops the performance of the \texttt{reacher-four} task. 
This result reveals that embodiments with similar dynamics or morphological features can facilitate more effective knowledge transfer during meta-testing, suggesting that the diversity of data greatly impacts performance.

\clearpage

\end{document}